\documentclass[fleqn]{wlscirep}
\usepackage{anyfontsize}
\usepackage[utf8]{inputenc}
\usepackage[T1]{fontenc}
\usepackage{enumitem}

\usepackage[left,mathlines]{lineno}
\usepackage{graphicx, amsmath, amsfonts, amsthm} 

\usepackage{array}
\usepackage{booktabs}
\usepackage{ragged2e}
\usepackage{makecell}

\newcolumntype{C}[1]{>{\centering\arraybackslash}m{#1}}
\newcolumntype{L}[1]{>{\RaggedRight\arraybackslash}m{#1}}

\newtheorem{prop}{Proposition}

\graphicspath{{figures/}} 

\title{Designing probabilistic AI monsoon forecasts to inform agricultural decision-making}

\date{January 2026}
\author[1,*]{Colin Aitken}
\author[2]{Rajat Masiwal}
\author[2]{Adam Marchakitus}
\author[3]{Katherine Kowal}
\author[4]{Mayank Gupta}
\author[5]{Tyler Yang}
\author[7]{Amir Jina}
\author[2,3]{Pedram Hassanzadeh}
\author[5,8]{William R. Boos}
\author[1,6,*]{Michael Kremer}

\affil[1]{Development Innovation Lab, University of Chicago, IL, 60637}
\affil[2]{Department of the Geophysical Sciences, University of Chicago, IL, 60637}
\affil[3]{Data Science Institute, University of Chicago, IL, 60637}
\affil[4]{Development Innovation Lab India, University of Chicago Trust, India, 560025}
\affil[5]{Department of Earth and Planetary Science, University of California, Berkeley, California, 94720}
\affil[6]{Kenneth C. Griffin Department of Economics, University of Chicago, IL, 60637}
\affil[7]{Harris School of Public Policy, University of Chicago, IL, 60637}
\affil[8]{Climate and Ecosystem Sciences Division, Lawrence Berkeley National Laboratory, Berkeley, California, 94720}
\affil[*]{Corresponding authors: caitken@uchicago.edu, kremermr@uchicago.edu}

\begin{abstract} 
Hundreds of millions of farmers make high-stakes decisions under uncertainty about future weather\cite{davis2023estimating, burlig2024_value_of_forecasts}. Forecasts can inform these decisions\cite{gine2017forecasting, rosenzweig2019assessing, rudder2024learning, yegbemey2021_sms_maize, fosu2018disseminating}, but available choices and their risks and benefits vary between farmers\cite{mulungu2024one, kuivanen2016characterising, akzar2023understanding, krishna2022gender}. We introduce a decision-theory framework\cite{blackwell1953equivalent,epstein1962_bayesian_meteorology, murphy1977_value_of_forecasts} for designing useful forecasts in settings where the forecaster cannot prescribe optimal actions because farmers’ circumstances are heterogeneous.   We apply this framework to the case of seasonal onset of monsoon rains, a key date for planting decisions and agricultural investments in many tropical countries. We develop a system for tailoring forecasts to the requirements of this framework by blending systematically benchmarked\cite{masiwal2025monsoonAI}  artificial intelligence (AI) weather prediction models with a new “evolving farmer expectations” statistical model. This statistical model applies Bayesian inference to historical observations to predict time-varying probabilities of first-occurrence events throughout a season.  The blended system yields more skillful Indian monsoon forecasts at longer lead times than its components or any multi-model average. In 2025, this system was deployed operationally in a government-led program that delivered subseasonal monsoon onset forecasts to 38 million Indian farmers, skillfully predicting that year’s early-summer anomalous dry period\cite{vallangimonsoon25}. This decision-theory framework and blending system offer a pathway for developing climate adaptation tools for large vulnerable populations around the world.

\end{abstract}

\begin{document}

\flushbottom
\maketitle

\thispagestyle{empty}

\section*{Introduction}
Farmers worldwide make high-stakes decisions about planting, crop choice, and input investments under considerable uncertainty about future weather \cite{davis2023estimating, hansen2011review}. Randomized experiments find that exposure to weather forecasts meaningfully shapes these decisions\cite{burlig2024_value_of_forecasts, fosu2018disseminating, yegbemey2021_sms_maize, rudder2024learning}. In tropical regions, for example, key decisions hinge on the timing of rainy-season onset: crops may fail if planted before dry spells transition to continuous rain, while expectations of rainy season length shape crop choice and investment intensity \cite{ati2002comparison, wakjira2021rainfall, gadgil2006indian}.  However, many smallholder farmers lack access to forecasts of key weather events, particularly at longer lead times\cite{yegbemey2023weather, tamru2025climate, linsenmeieretal}. 
Recent advances in artificial intelligence weather prediction (AIWP) offer an opportunity to close this gap, with AIWP models matching or exceeding the skill of state-of-the-art numerical weather prediction (NWP) models \cite{allen2025end, ben2024rise,bi2023accurate,lam2022graphcast,pathak2022fourcastnet,price2025probabilistic, nath2026predicting} at a fraction of the computational cost. Many AIWP models are also open-source and easy to use and tailor. 
If AIWP forecasts can be translated into comprehensible and actionable information, these advances could deliver substantial benefit to hundreds of millions of smallholder farmers.

Realizing these benefits, however---whether through AIWP or NWP or a blend of both---requires confronting a fundamental design problem.  Although weather forecasts can help farmers improve decisions \cite{gine2017forecasting, rosenzweig2019assessing}, the appropriate agricultural response to a forecast may depend on information the forecaster does not observe, including individual farmers’ land characteristics, access to inputs, and experience with different crops and technology. These attributes can be heterogeneous across farmers\cite{hansen2011review, mulungu2024one, kuivanen2016characterising, akzar2023understanding, krishna2022gender}. For example, a risk-averse farmer may require greater certainty before planting, even if waiting reduces expected yields; a farmer with access to irrigation or drought-tolerant seed varieties may be more willing to plant before a potential dry spell after initial rains. Similarly, a farm household with a member with outside employment may be more willing to take risks than one fully dependent on farm income. This may make it infeasible for central authorities to prescribe optimal agricultural decisions for each farmer. We address the question of how to design forecasts to empower farmers to reach their own decisions.


The value of a forecast also depends on how it updates farmers’ expectations relative to what they would have believed without the forecast. Many forecasts are conventionally evaluated against a static ``climatology” baseline consisting of the historical median onset date at a location \cite{bombardi2017sub, goswami2010evaluation, scheuerer2024probabilistic}. In a number of important practical cases, like rainy season onset, such a baseline can be highly unrealistic from the perspective of a farmer making agricultural decisions. Consider a year in which onset is delayed; when a static climatology forecast is examined after the historical median onset date but before the actual onset has occurred, that static forecast would predict that onset has likely already happened. In contrast, a farmer would be aware that continuous rains have not yet started and would expect onset to occur in the future. The static nature of the baseline means that using it in a benchmark will exaggerate a forecasting model's value to farmers, whose beliefs about likely onset dates evolve throughout the season.

We address these problems here, leveraging decision theory to guide forecast design in settings where end users have varying objectives, constraints, and risk tolerances. We develop an ``evolving-expectations'' statistical model in which the distribution of future onset probabilities is dynamically updated throughout the season to reflect the fact that, as the season progresses, a farmer will observe for some period of time that onset has \emph{not} yet occurred. 
We then combine the evolving-expectations model with AIWP models, selected using a decision-oriented benchmark \cite{masiwal2025monsoonAI}, to construct blended {\it probabilistic} statistical–AI forecasts of local Indian monsoon onset. 
This blending not only combines information from multiple forecast models found to be individually skillful and operationally practical \cite{masiwal2025monsoonAI}, but dynamically weights the contributions of its components by lead time.  This makes the blending approach more appropriate than a traditional multimodel ensemble for a setting in which AIWP or NWP model skill declines more quickly with lead time than that of the evolving-expectations model. 

This work builds on evidence that farmers can use information about local rainy season onset to adjust input and planting decisions \cite{burlig2024_value_of_forecasts}. Many of these decisions benefit from predictions with lead times longer than one week \cite{gine2017forecasting}, but publicly available onset forecasts either provide a single date for the entire country or are available only at shorter lead times. The India Meteorological Department (IMD) releases a skillful forecast with multiple weeks of lead time for monsoon onset over Kerala, in southern India \cite{pai2009summer, pattanaik2024objective}, but this has little correlation with rainy season onset dates in the rest of the country \cite{moron2014interannual}. The IMD also releases maps denoting the current extent of monsoon progression, as well as short-term forecasts of progression over the next 1-3 days. Farmers thus lack local predictions of monsoon onset at lead times that can help with longer-term planning or planting decisions. 

The framework and models presented here and in Masiwal {\it et al.}~\cite{masiwal2025monsoonAI} were developed to fill this gap, and were used in a 2025 program of India's Ministry of Agriculture and Farmers' Welfare that distributed probabilistic local onset forecasts weekly to 38 million farmers across 13 states\cite{vallangimonsoon25}.


\section*{Decision-Theory Framework Informing Forecast Design}
 
 We use the tools of decision theory\cite{epstein1962_bayesian_meteorology, murphy1977_value_of_forecasts}, starting with Blackwell's Informativeness Theorem \cite{blackwell1953equivalent},  to develop a model to guide a forecaster choosing a  weather prediction to send to a heterogeneous group of farmers. The farmers make agricultural decisions whose payoffs depend both on future weather and additional information known to each farmer but not to the forecaster. The forecaster obtains signals about future weather from prediction models. These assumptions allow derivation of three key considerations for forecast design and evaluation, summarized in Table \ref{table:implications}.

First, probabilistic forecasts allow heterogeneous farmers to tailor decisions to their constraints in ways that deterministic forecasts or advisories do not. We show that a well-calibrated probabilistic forecast (i.e., with probabilities aligning with empirical frequencies) leads to better outcomes than a deterministic message built from that forecast (Proposition \ref{probabilistic}, Supplementary Fig. S1)\cite{krzysztofowicz2001_probabilistic_hydrology, murphy1977_value_of_forecasts}. For example, a farmer whose livelihood is at stake if crops fail may consider a 30\% chance of the rainy season's false onset too risky to plant, while planting early may pose less of a risk to a farmer with secure income outside of agriculture. A deterministic forecast that flattens the probability into `` true onset is not expected in the next 2 weeks" will not meet both farmers' informational needs. More generally, the decision-theory model implies that forecasters should avoid ``coarsening'' messages by removing potentially relevant information. In practice, while end users are capable of interpreting probabilistic information, comprehension varies with phrasing and context.\cite{ripberger2022communicating, patt2005effects, luseno2003assessing}.  Careful testing is therefore essential to arrive at appropriate message framing and text. 

Second, forecasts will more robustly benefit farmers if they incorporate information to which farmers already have access. Some farmers may perfectly incorporate new forecasts with their existing knowledge, and they will benefit as long as the forecasts contain new information. However, other farmers may take forecasts at face value, and they will only benefit if the forecast is better than their priors, i.e., information they would have used in the absence of forecasts.  A corollary is that forecasts should be evaluated relative to a baseline containing relevant information already available to both forecasters and farmers. For forecasts of an event's timing, e.g., monsoon onset, this motivates the “evolving-expectations” statistical model presented in the next section. Baselines that do not represent farmer expectations can substantially overstate the value of new predictive models and lead to dissemination of information worse than what farmers already have.  To guarantee farmers are better off in expectation, forecasts need to fully incorporate farmers' existing information (Proposition \ref{benefit}). In practice,  incorporating every farmer's private information about the weather would be infeasible, but forecasters can approximate it by blending relevant, publicly available information into forecast models. 


Third, the decision-theory model implies that as long as farmers correctly understand the forecast, it is possible to identify whether the forecast benefits farmers in expectation by testing whether farmers change their decisions. This motivates measuring forecast value by whether individuals change their decisions in response to information, rather than on whether researchers deem those decisions to be “correct.” When farmers make decisions to manage potential risks, an evaluation of forecast impacts on yields or profits in a single season cannot, in general, determine whether the forecast was ex-ante beneficial. This is because the evaluator will not observe the full distribution of potential weather events and other variables that affect these outcomes each season (Proposition \ref{no-outcome-rct}). Indeed, risk-averse farmers  may respond to accurate forecasts in ways that make them better off but reduce expected profits (Proposition \ref{manage-risk}).  For example, a farmer may reasonably hedge against a potential dry event by purchasing drought-tolerant seeds with lower expected yields. Observing that yields were lower in response to forecasts will not allow the evaluator to determine whether forecasts helped the farmer manage their risk. 

A valid test of forecast relevance does not then depend on researchers' knowledge of best practices for local farming or measures of impact on variables such as yield or farmer profits (Proposition \ref{sufficient}). Changes in decisions can generally be measured more precisely than changes in outcomes---such as crop yield or income, which are affected by factors beyond farmers' decisions---and therefore are practical for testing forecast relevance when sample sizes are limited (Proposition \ref{decision-power}).

\begin{table}[h]
\centering
\renewcommand{\arraystretch}{1.3} 

\begin{tabular}{|L{7cm}|L{8cm}|}
\hline
\multicolumn{1}{|C{7cm}|}{\textbf{Model Result}} &
\multicolumn{1}{C{8cm}|}{\textbf{Implication for Forecast Design and Evaluation}} \\
\hline

Probabilistic forecasts better allow heterogeneous farmers to make their own decisions based on information relevant to farming that the forecaster does not have (Proposition \ref{probabilistic}).
&
\begin{itemize}[leftmargin=*, topsep=6pt, itemsep=2pt, parsep=0pt, after={\vspace{0pt}}]
\item Provided probabilistic forecasts.
\item Designed messages in part based on focus groups with farmers aimed at increasing comprehension of probabilistic information.
\end{itemize}
\\
\hline

If some farmers take forecasts at face value, guarantees on benefits of forecasts require forecasts to be well-calibrated and incorporate farmers' other sources of information about weather outcomes (Proposition \ref{benefit}).
&
\begin{itemize}[leftmargin=*, topsep=6pt, itemsep=2pt, parsep=0pt, after={\vspace{0pt}}]
\item Designed a new ``evolving expectations" statistical model as both a baseline and a component of the blended model.
\item Used the blended model, which met the criteria for dissemination, rather than AIWP model outputs on their own, which did not.
\item Tested calibration of blended model across multiple time periods.
\end{itemize}
\\
\hline

Experiments measuring yield and profit outcomes do not capture the value of forecasts if farmers are risk-averse. Whether forecasts have value to farmers can be measured by observing whether farmers change decisions in response to forecasts (Propositions \ref{no-outcome-rct} - \ref{decision-power}).
&
\begin{itemize}[leftmargin=*, topsep=6pt, itemsep=2pt, parsep=0pt, after={\vspace{0pt}}]
\item Identified monsoon onset as valuable target based on previous research finding that farmers change decisions after receiving monsoon onset forecasts \cite{burlig2024_value_of_forecasts}. 
\item Chose an agriculturally-relevant monsoon onset definition, capturing the start of continuous local rainfall not followed by a potentially crop-damaging dry spell, to specifically target farmer decisions. 

\end{itemize}
\\
\hline

\end{tabular}
\caption{Implications of decision-theory framework and operational decisions informed by the results, and actions we took as a result.}
\label{table:implications}
\end{table}

\section*{An Evolving-Expectations Model}

\begin{figure}[htbp]
  \centering
  \includegraphics[width=\textwidth]{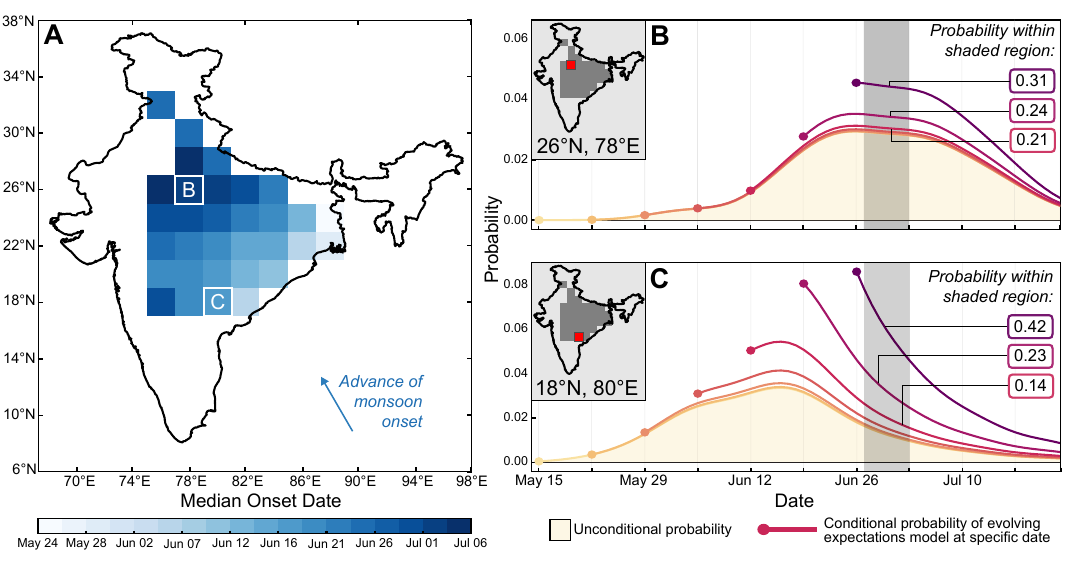}
\caption{\textbf{Static climatology and the evolving-expectations model.} \textbf{A)} Median onset date for the grid cells used in dissemination. \textbf{B)} The distribution of possible onset dates predicted by the evolving-expectations model for a grid cell centered at latitude 26$^\circ$N and longitude 78$^\circ$E. Shaded distribution shows the unconditional probability of onset; lines show the probability from the evolving expectations model if the onset has not occurred by the date signified by the dot. Illustrative probabilities are calculated over the shaded grey region. As the season progresses without an onset, the probability of onset in any particular future week increases. \textbf{C)}  The distribution of possible onset dates predicted by the evolving-expectations model for a grid cell centered at 18$^\circ$N and 80$^\circ$E. The decision-theory framework implies that models that cannot outperform this baseline should not be used on their own for dissemination. } 
  \label{fig:fig1}
\end{figure}

The decision-theory framework establishes that forecasts should be evaluated relative to a baseline (i.e., reference forecast) that includes a model of farmers' prior beliefs in the absence of forecasts. For monsoon onset, this means the baseline should include a simple but powerful piece of information: whether or not rains have already started locally. Following the framework's implications for forecast design, we use an existing, well-studied agronomic definition of onset as the first wet spell of the season that is not soon followed by a prolonged dry period that could prove harmful to crops\cite{moron2014interannual} (see Methods). We modify the conventional climatology predictor (i.e., median or mean historical onset date) by developing an ``evolving-expectations'' statistical model that applies Bayesian updating to the historical distribution of onset dates, calculated using IMD rain-gauge data, conditioning on whether onset has occurred in a given season. As the season progresses with onset unobserved, the model shifts probability mass forward in time, concentrating it in the remaining plausible onset weeks (Fig.~\ref{fig:fig1}).

This modified baseline has significant implications. For example, for a region around $18^\circ$N (Fig.~\ref{fig:fig1}C), if onset has not occurred by June 26, the static (unconditional) climatology predicts the probability of onset  to be $9\%$ in the following week, while the evolving-expectations model predicts a $42\%$ probability. The decision-theory framework implies that forecasts should not be disseminated if they cannot outperform this new baseline, which captures information about onset that is available to farmers (Table~\ref{table:implications}).


We evaluate probabilistic forecasts of onset produced by this model using three complementary metrics: the Brier Score, which measures the accuracy of predicted probabilities for onset in specific weeks; the Ranked Probability Score (RPS), which additionally accounts for how close a predicted week is to the true onset date; and the area under the receiver operating characteristic (ROC) curve (AUC), which measures the ability to discriminate between likely and unlikely onset weeks irrespective of the model's calibration (see Methods). 
We primarily assess the 2000–2024 cross-validation period, but also include results from an additional 1965–1978 hold-out period, noting that initial conditions used in the AIWP models may be less accurate in that pre-satellite era\cite{masiwal2025monsoonAI}.

\begin{figure}[htbp]
  \centering
  \includegraphics[width=\textwidth]{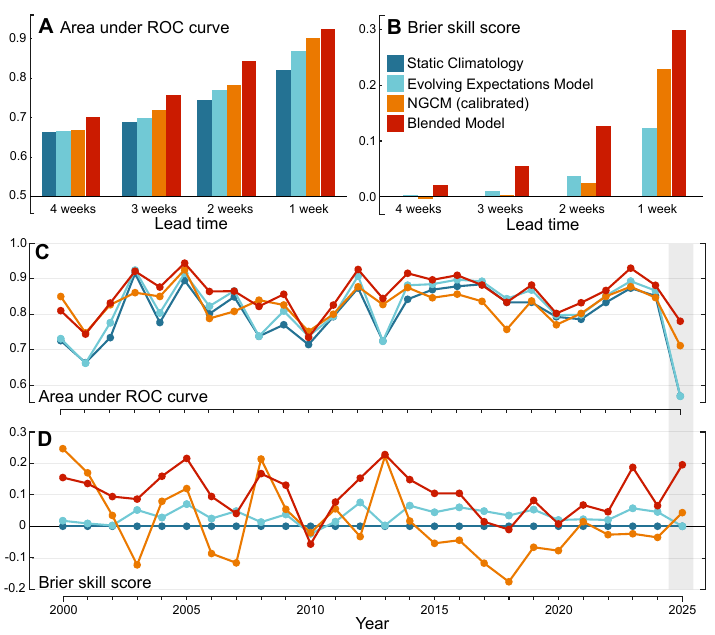}
\caption{\textbf{Evaluation of temporal components of models' skill for the 2000-2024 period.} \textbf{A)} Area under ROC curve (AUC) by lead time. The baseline for the bars is chosen to be 0.5, indicating the AUC of a forecast with no ability to distinguish onsets from non-onsets. \textbf{B)} Brier skill score by lead time, computed relative to a traditional (static) climatology model. \textbf{C)} AUC by year across all lead times. The 2000-2024 scores are computed via cross-validation. The 2025 scores are only for forecasts that were actually disseminated before Moron-Robertson onset in each grid cell. Dissemination began in late May, so the set of initialization dates is smaller than in other years. \textbf{D)} Brier skill score by year across all lead times, calculated relative to a traditional (static) climatology model. A deterministic version of AIFS was used in both the blended model and benchmarking, so NGCM is shown here as a representative AIWP model as it performs better on these probabilistic metrics (Fig.~\ref{fig:fig4}). See Extended Data Fig.~2 for results for the 1965-1978 period.} 
  \label{fig:fig2}
\end{figure}

Figure \ref{fig:fig2} shows the AUC and Brier Skill Score (BSS) for static climatology, the evolving expectations model, a calibrated hybrid AI model (NGCM; calibration is described below), and a model that blends two AI models and the evolving-expectations model (described in the next section). The evolving-expectations model outperforms static climatology at all individual weekly lead times (Fig. \ref{fig:fig2}A, B) and in both validation periods (Extended Data Figs. 1, 2) for all metrics. It outperforms raw (i.e., uncalibrated) probabilistic forecasts from a hybrid AI model (NGCM) shown to be skillful on its own for monsoon onset predictions\cite{masiwal2025monsoonAI} (Fig.~\ref{fig:fig4}). Raw NGCM predictions are overconfident when comparing observed frequencies with predicted probabilities (Fig.~\ref{fig:fig3}A), but the evolving-expectations model even outperforms the calibrated NGCM for some metrics at longer lead times (Fig. \ref{fig:fig2}B) and when all lead times out to four weeks are considered jointly (Extended Data Fig. 1).

At these longer lead times, the AIWP models on their own therefore do not meet the second criterion outlined in the decision-theory framework for dissemination, which requires that forecasts improve on a farmer's prior beliefs. This motivates a blended approach that combines both information sources.

\section*{Blending AI and Statistical Forecasts}

We create a blended model (see Methods) that combines probabilistic information from the evolving-expectations model with rainfall forecasts from two AIWP models selected as part of a decision-relevant benchmarking\cite{masiwal2025monsoonAI}: Google's NGCM\cite{kochkov2024neural} and the European Centre for Medium-Range Weather Forecasts' AIFS\cite{lang2024aifs}. AIFS is a purely data-driven model, while NGCM is a hybrid model combining differentiable dynamical core with neural network parameterizations of physical processes.  

\begin{figure}[htbp]
  \centering
  \includegraphics[width=\textwidth]{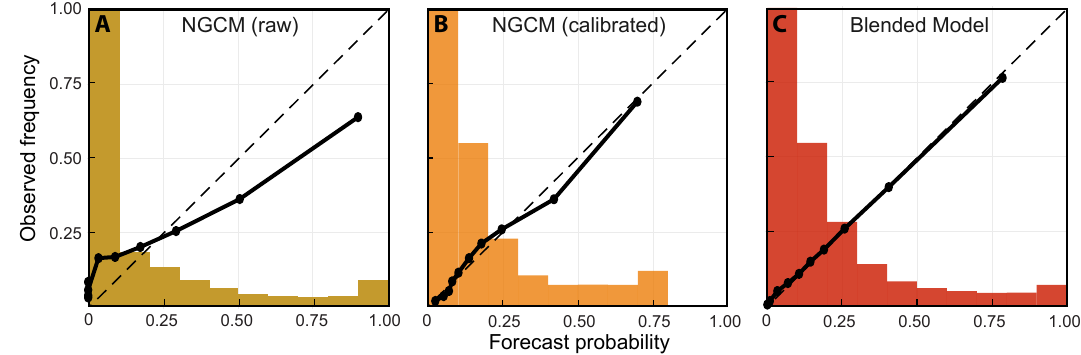}
\caption{\textbf{Reliability diagram and histogram of probabilities assigned by different models during the 2000-2024 cross-validation period.} If the model is well-calibrated the points should line up with the dashed line.  Each point represents a decile of probabilities assigned by the model, and compares the average predicted probability in that decile and the fraction of events which occurred in observation. Histograms are normalized so that the bin capturing probabilities between 0 and 10\% has height equal to one. A deterministic version of AIFS was used for both benchmarking and the blended model, so NGCM is used as a representative AIWP model for this figure. See Extended Data Fig.~3 for the results from the 1965-1978 period.}
  \label{fig:fig3}
\end{figure}

\begin{figure}[htbp]
  \centering
  \includegraphics[width=\textwidth]{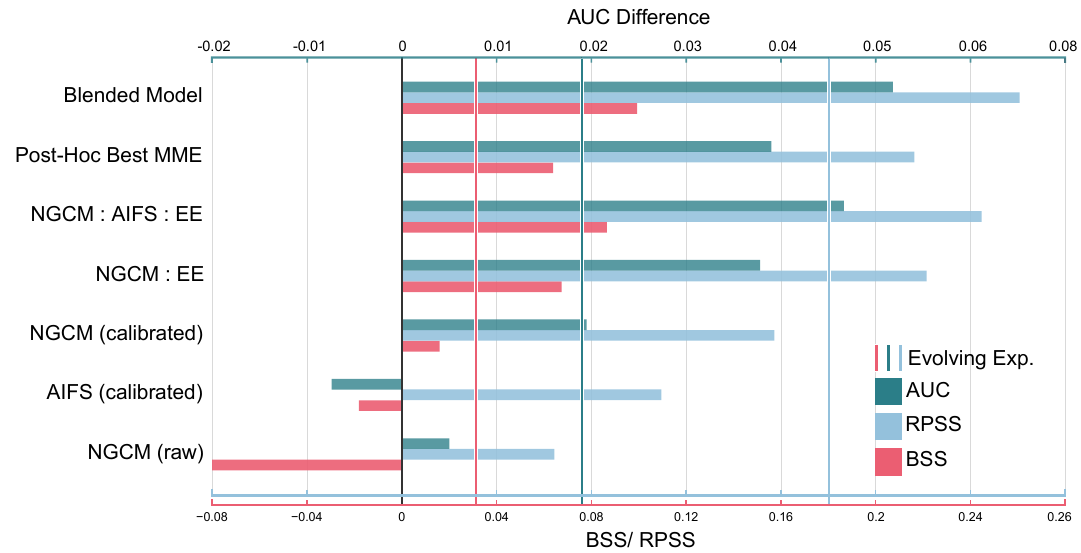}
  \caption{\textbf{Performance of models during the primary 2000-2024 period, including submodels of the blended model.} Vertical lines represent the evolving expectations model, which the decision-theory model implies should be a baseline for dissemination. The interaction models, indicated with colons, are submodels of the final blended model combining the evolving expectations model (EE) with 5-day rainfall variables from one or more AIWP models but not 10-day rainfall variables.  The best multimodel ensemble (MME) is selected as a linear combination of probabilities from AIFS, NGCM, and the evolving expectations model maximizing the RPSS post-hoc on the 2000-2024 period. All other models (including calibration) are trained via leave-one-year-out cross-validation. Skill scores and differences in AUC are computed relative to unconditional climatology, which has an AUC of .810, a Brier Score of 0.583, and a RPS of 0.611. See Extended Data Fig.~4 for the results from the 1965-1978 period.}
  \label{fig:fig4}
\end{figure}

Intuitively, the blended model upweights AIWP-predicted wet spells when they are likely to correspond to true onset and downweights them on calendar dates when they are more likely to be a false onset (i.e., followed by a prolonged dry spell). The relative influence of each information source can vary with lead time: a model’s prediction of rainfall in the coming week may receive greater weight than climatology, even as its prediction of dry conditions four weeks ahead receives less weight than climatology. The flexible weighting allows each information source to contribute most strongly where it is most informative, which is not achieved by traditional multimodel ensembles or simple averaging methods. The blended model's multinomial logistic regression specification also ``calibrates'' the probabilistic forecasts (see reliability diagrams in Fig.~\ref{fig:fig3} and Extended Data Fig.~3). 

The blended model outperforms both the static (unconditional) climatology baseline and the evolving-expectations (conditional climatology) baseline across all three metrics and all evaluation periods, including the cross-validation model selection period (2000–2024; see next paragraph), the pre-satellite-era hold-out period (1965–1978), and the 2025 dissemination period (Fig.~\ref{fig:fig2} and Extended Data Figs. 1-2). Note that the 2025 comparison uses operational forecasts produced for dissemination and publicly archived on creation (see Data Availability section).

Cross-validation produces forecasts for each year by training the model on all other years during the period, ensuring that predictions are scored on held-out data.  When pooled across lead times up to four weeks, the blended model yields a 5–10\% improvement in Brier Score, a 20–25\% improvement in RPS, and a 3–5 percentage point increase in AUC relative to the static climatology baseline.  In 2025, when climatology performed poorly due to an unusual pause in the northward progression of monsoon onset~\cite{masiwal2025monsoonAI}, the blended model's forecasts yield a $\sim20\%$ improvement in Brier Score and RPS as well as a $20$ point increase in AUC relative to the static climatological baseline. 

The blended model's skill is highest at a one-week lead time, showing roughly a 15\% improvement in Brier Score relative to the evolving-expectations baseline (25\% relative to static climatology) during 2000–2024, and declines gradually but remains positive out to four weeks (Fig. \ref{fig:fig2}). Skill patterns are qualitatively similar but smaller in magnitude in the 1965–1978 hold-out period (Extended Data Fig. 2), with improvements over the evolving-expectations model persisting to three-week lead times. Forecasts remain well calibrated in both the cross-validation and hold-out periods (Fig. \ref{fig:fig3} and Extended Data Fig. 3) and skill gains occur over a wide geographic region (Supplementary  Figs. S2-S4).

The blended model does more than just combine and calibrate forecasts from different models. Forecasting the dry spell criterion in the agronomic onset definition requires a forecast that is 30 days longer than the lead time; e.g., 45-day and 60-day rainfall forecasts are needed for onset predictions with 15-day and 30-day lead times, respectively (Methods). Current AIWP models do not have such long-range skill for rainfall~\cite{masiwal2025monsoonAI} and cannot be used to directly forecast the full agronomic onset, with its dry spell criterion. Blending with the evolving-expectations model addresses this problem by weighting AIWP rainfall forecasts according to the climatological probability of onset. Intuitively, an AIWP prediction of a five-day wet spell gets more weight when that prediction is more likely to reflect a true onset, and less weight when onset is unlikely and hence the predicted five-day wet spell might by followed by a dry period. In addition, since rainfall variables are included as continuous variables, the blended model implicitly performs simple bias correction, adjusting the five- and ten-day cumulative rainfall forecasts by a different constant for each week of lead time. 

The evolving expectations model is trained on 124 years of Indian rain-gauge data \cite{pai2014_india_gridded_rainfall}, raising the question of whether it could be used in other monsoon regions with more limited data availability (e.g., East Africa). Although the statistical model alone loses skill when trained on fewer years of data, these losses do not meaningfully propagate to the blended model, indicating that the approach remains viable under data sparsity (Extended Data Fig. 5). This may be because the AIWP forecasts and climatology have implicit information in common, so that the blended model's structure allows other inputs to compensate when the quality of one declines.

The value of the blending approach 
is further explored by comparing it with a standard multimodel ensemble approach, often used as a default for combining NWP model predictions\cite{scheuerer2024probabilistic, vannitsem2021_stat_postprocessing_review}.  The predicted probabilities of the multimodel ensemble are taken to be a weighted average of those implied by individual models. Special cases of this approach, such as Bayesian model averaging, correspond to different methods of choosing the weights. During the 2000–2024 cross-validation period, the blended model outperforms every fixed-weight ensemble, even when ensemble weights are selected \emph{ex post} (Fig.~\ref{fig:fig4}). In the 1965–1978 hold-out period, this remains true for two of the three metrics, although the best ensemble achieves a slightly higher RPSS (Extended Data Fig.~4).

\section*{Discussion}
The AI revolution has the potential to change hundreds of millions of lives by providing skillful forecasts of consequential weather phenomena. However, onset forecasts from these AIWP models do not, on their own, meet the criteria our decision-theory model gives for dissemination at lead times past one week (Table~1 and Fig.~\ref{fig:fig2}). We instead satisfy these criteria by combining multiple AIWP model forecasts with a model of farmers' evolving-expectations based on historical rain-gauge data (Fig.~\ref{fig:fig1}). While this framework can also use NWP forecasts, AIWP models offer key practical advantages. In addition to having accuracy comparable to or better than that of NWP models, AIWP models can more easily produce large hindcast ensembles with the same model specification used for real-time operation.  This facilitates decision-oriented benchmarking and addresses the small test-sample size problem for infrequent phenomena such as monsoon onset\cite{masiwal2025monsoonAI}, and also enables a blending approach that leverages the individual strengths of multiple models. Forecasters can improve skill for particular phenomena by adding existing models to a blend even when they cannot improve or tailor the underlying weather models themselves.   


To illustrate the magnitude of the skill improvement provided by the blended model, we place our results within the broader forecasting literature while noting that these are not direct comparisons. At one-week lead times, the blended model's BSS of 30\% is broadly comparable to skill scores reported for the Global Ensemble Forecast System (GEFS)\cite{zhou2022development_gefs} and the European Centre for Medium-Range Weather Forecasts (ECMWF ensemble)\cite{ecmwf_brier_precip} precipitation forecasts at 4-5 day lead times. This is representative of the upper range of skill for major operational forecasts of precipitation-related metrics over the continental United States and Europe/North Africa respectively. 
Also in the first week, the evolving-expectations model improves on static climatology by about 10 percentage points and the blended model improves on calibrated NGCM forecasts by about 7 percentage points. These are roughly comparable to the increase in BSS between GEFS version 11 and GEFS version 12 precipitation forecasts at up to one-week lead times, a major development\cite{zhou2022development_gefs}. Skill necessarily declines with lead time; however, the positive skill out to four weeks is noteworthy in the context of subseasonal-to-seasonal forecasting where state-of-the-art AIWP and NWP models show near-zero or negative BSS for weekly precipitation at weeks 3–4 \cite{chen2024machine}.

The blended model had large skill in 2025 even though that year's monsoon onset was atypical (Fig. \ref{fig:fig2}). The monsoon first reached mainland India eight days earlier than normal on May 24, but by May 29 its northward progression stalled for more than two weeks. This unusual progression meant the climatological baseline performed poorly, but the blended model's accuracy in 2025 was similar to its accuracy in a typical year, so its skill relative to the climatological baseline was high. 

Most existing monsoon onset forecasts are produced once at the start of the season\cite{scheuerer2024probabilistic, pai2009summer}. In contrast, weekly updating allows forecast skill to increase as the true onset date approaches, enabling users to incorporate progressively more informative guidance throughout the planning cycle. Forecasts of monsoon onset and other periodic phenomena (such as first frost or the first hurricane of a season) should consider adopting evolving climatological baselines, constructed by conditioning historical distributions on the event not having occurred by the forecast date. Similar principles might apply to other quasi-periodic climate phenomena, including the El Niño–Southern Oscillation and the Madden–Julian Oscillation. More generally, forecast evaluation should employ baselines that include information already available to users, ensuring that reported skill corresponds to true added value (Table~1).


One practical limitation of the 2025 implementation was that forecasts were produced at weekly resolution. This arose from both the multinomial logit functional form, in which the number of coefficients scales quadratically with the number of forecast bins, and the variability of onset timing within a grid cell. Future work might retain skill gains from the blending procedure while enabling daily forecast resolution, and also potentially enhance spatial resolution finer than $2^\circ$. Additional models (including NWP models) and dynamical information could be added to the blended framework. 

The best models and predictors to include in a blended framework may differ from those that perform best in benchmarking exercises on their own, as the relevant question is which models and variables provide the most information \emph{not already contained} in other models' forecasts\cite{kang2022forecast}.
Traditional predictor selection procedures, including sparsity-promoting techniques such as LASSO\cite{tibshirani1996regression}, may be further effective in conjunction with cross-validation to avoid overfitting.  In this case, the main role of decision-oriented benchmarking\cite{masiwal2025monsoonAI} in a blending procedure may be to identify models with flaws that would make them unsuitable for inclusion, rather than to select the best models to be blended. 

Realizing the potential benefits of AIWP for applications such as farming requires more than building accurate weather prediction models: it requires identifying knowledge that decision-makers may already have, and designing statistical frameworks to produce decision-relevant information by flexibly combining multiple sources of data. It also requires careful message design to ensure users can understand and make use of this information. The principles developed here offer a replicable pathway for building climate adaptation tools tailored to large, vulnerable populations across the tropics and beyond.


\section*{Methods}

The blended model is designed to predict the probability that onset in each $2^\circ \times 2^\circ$ grid cell occurs in each of four weeks following forecast initialization. A multinomial logistic regression specification combining probabilistic information from the evolving-expectations model with AIWP rainfall forecasts was evaluated using leave-one-year-out cross-validation over 2000–2024 using three probabilistic metrics: Brier Skill Score (BSS), area under the Receiver Operating Characteristic curve (AUC), and Ranked Probability Skill Score (RPSS). This procedure mitigates overfitting by evaluating models only on data they were not specifically trained on.
\subsection*{Data}
Daily rainfall data from the India Meteorological Department (IMD) observational rain-gauge network\cite{pai2014_india_gridded_rainfall} regridded to a $2^\circ$ by $2^\circ$ horizontal resolution was used as both a ground truth data source and an input into the climatology models. Precipitation forecasts from NGCM (30 ensemble members) and AIFS were regridded to the same resolution and initialized twice weekly, beginning in May of each year through the date of monsoon onset for each grid cell. Overall, the model was trained on 32 grid cells selected for potential dissemination (shown in Fig.~\ref{fig:fig1}).

Grid cells were selected for dissemination based on several factors, including the skill of the underlying AIWP weather models, the climatological likelihood of dry spells following an initial five day wet spell, and the variability of the Moron-Robertson onset date within the two degree grid cells.

\subsection*{Onset Definition}
Forecasts were for a modified Moron-Robertson \cite{moron2014interannual} definition of monsoon onset for a $2^\circ \times 2^\circ$ grid cell, which defines local monsoon onset as the first wet day ($\geq 1\mathrm{mm}$) of the first 5-day wet sequence after April 1 whose cumulative rainfall is at least the amount of the local climatological 5-day wet spell, which is not followed by any 10-day dry spell ($<5\ \mathrm{mm}$ in total) within the subsequent 30 days. This precipitation-based onset definition is more appropriate than circulation-based indices for agricultural applications.\cite{burlig2024_value_of_forecasts}  Focus groups indicated that farmers generally did not consider April or May rain to be monsoonal, even if the following dry spells did not meet the Moron-Robertson threshold. As a result, forecasts of very early triggers of the Moron-Robertson onset definition would be unlikely to affect farmer decision-making and hence have limited value. We therefore further restricted our onset definition to consider dates after the monsoon reaches Kerala, the first part of India to receive the monsoon, as declared by the India Meteorological Department (IMD)\cite{pai2009summer, joseph2006summer}.

\subsection*{Climatology and Evolving-Expectations Model Specifications}\label{climdesc}
 
 The evolving-expectations model is used both as a baseline for benchmarking and an input into the blending algorithm. To avoid small-sample problems creating instability during periods where onset is less likely, a Kernel Density Estimator was fit to convert past onset dates for each grid cell into a smooth probability distribution. 

The Kernel Density Estimator (with Gaussian kernel) estimates the probability distribution as a mixture of normal distributions, with one normal distribution in the mixture centered at each previous onset date. In other words, if a grid cell has observed onset dates $d_1, d_2, \cdots, d_n$ over the past $n$ years, the Kernel Density Estimate of the probability density function for onset dates is:
\[
f(x) = \frac 1n \sum_{i=1}^n N^\sigma(x - d_i),
\]
where $N^\sigma(x)$ denotes the probability distribution of a normal distribution with mean 0 and standard deviation $\sigma.$ Here $\sigma$ is selected via the Sheather-Jones method \cite{sheather1991reliable}, a data-driven approach designed to minimize the out-of-sample mean integrated squared error of the estimator.

We call the resulting probability distribution (static or unconditional) \emph{climatology}, for the given grid cell, and use it as a baseline comparator. As input into the blended model and an additional baseline, we also produced \emph{evolving-expectations} probability distributions for each grid cell and each forecast date, which is constructed by conditioning the climatological distribution on the fact that onset has not occurred by the forecast initialization date. Equivalently, the evolving expectations distribution is the Bayesian posterior for a farmer whose prior belief was the climatological distribution and has observed onset not to have occurred yet. This provides the model with additional information: for example, if onset usually occurs June 15 and the forecast is being made June 30, the model ``knows'' that onset may be likely to come very soon even if the unconditional probability of an onset in early July is quite low (Fig.~\ref{fig:fig1}).

\subsection*{Blended Model Specification}

The blended model uses a multinomial logit specification to predict whether onset will occur in the first, second, third, fourth, or after the fourth week following the forecast initialization date.  We refer to each of these time units as a ``bin''. For a given forecast $i$ (where $i$ incorporates both the grid cell and the initialization date) and bin $j$ (e.g. week 1, week 2, etc.), we define the following:
\begin{enumerate}
    \item $p_{ij}$ is the probability the evolving expectations model assigns to bin $j$, conditional on not having occurred at the time of the forecast, winsorized\footnote{For $a < b$, the winsorization of a variable $x$ to lie between $a$ and $b$ is defined as $\max(a, \min(x, b)).$} to lie between $.0001$ and $.99$. 
    \item $\pi_{ij} = \log \left(\frac{p_{ij}}{1-p_{ij}}\right)$ is the logit-transformed probability of onset in bin $j$. We perform this logit transformation so the probabilities are on a logit scale when being fed into the multinomial logit procedure. 
    \item $\alpha_{ij}$ and $\nu_{ij}$ are the maximum amount of rainfall predicted by AIFS and NGCM respectively in a five-day period beginning in bin $j$, minus the five-day onset threshold for the grid cell corresponding to forecast $i$. The NGCM rainfall prediction for each day is defined as the mean rainfall predicted across all ensemble members.
    \item $\beta_{ij}$ and $\mu_{ij}$ are the minimum amount of rainfall predicted by AIFS and NGCM respectively in a ten-day period beginning in bin $j$, computed analogously.
\end{enumerate}
Interaction terms are included between $\pi$, $\alpha$, and $\nu$ variables within each week to capture correlations between the various onset predictors. For example, a negative coefficient on some $\pi_{ij} \alpha_{ij}$ term indicates a positive correlation between AIFS onset predictions and climatological onset probabilities, which we correct for to avoid model overconfidence. 

Let $Y_{ij}$ indicate whether observed onset occurs in bin $j$ for forecast $i$. We estimate a multinomial logistic regression of $Y_{ij}$ on the following variables
\[
\pi_{ij} ,\,  \alpha_{ij} , \,\nu_{ij} , \,\pi_{ij} \alpha_{ij} ,\, \pi_{ij} \nu_{ij} , \,\alpha_{ij} \nu_{ij} ,\,\pi_{ij}\alpha_{ij}\nu_{ij} ,\, \beta_{ij} , \,\mu_{ij} 
\]
for $j = 1,2,3,4.$ Note that we allow information from any lead-week bin $j$ to affect probabilities for each outcome bin $j'$. 

More explicitly, we estimate coefficients $t_{\ell,j,j'}$ $\ell \in \{0,...,9\},\, j \in \{1,...,4\}, $ and $j' \in \{1,...,4\}$ such that
\begin{align*}
\log\left(\dfrac{P(Y_{ij'}  = 1)}{P(Y_{i5}  = 1)}\right) = \sum_{j = 1}^4 &t_{0jj'} + t_{1jj'}\pi_{ij} +  t_{2jj'}\alpha_{ij} + t_{3jj'}\nu_{ij} + t_{4jj'}\pi_{ij} \alpha_{ij} \\&+ t_{5jj'}\pi_{ij} \nu_{ij} + t_{6jj'}\alpha_{ij} \nu_{ij} + t_{7jj'}\pi_{ij}\alpha_{ij}\nu_{ij} \\ &+ t_{8jj'}\beta_{ij} + t_{9jj'}\mu_{ij} 
\end{align*}
These ratios, along with the condition that $\sum_{j' = 1}^5 P(Y_{ij'} = 1) = 1$ determine the probabilities the forecast assigns to onset occurring within each bin. 

The interactions are intended to utilize the models' strengths in predicting five-day wet spells, weighting these according to the underlying climatological onset probabilities to down-weight wet spells during times when dry spells are likely.  The ten-day rainfall sums are intended to model forecasts relevant to predicting dry spells in later weeks, so they are not interacted with the probability of onset as the week of the dry spell is not likely to be the same as the week of onset.

\subsection*{Potential Sources of Bias}
This modeling procedure is subject to several potential sources of bias. First, the years 2000-2018 overlap with the training period for both AIWP models (with AIFS' training and fine-tuning extending through 2022), which could overstate their apparent accuracy and lead the blended model to put too much weight on them. However, this overlap is not necessarily a significant source of bias, as the Indian monsoon onset was not an explicit target of the global models, and when benchmarking raw individual model outputs we did not find evidence that AIWP models performed significantly better during the training period than testing period \cite{masiwal2025monsoonAI}. 

Second, because the AIWP models were selected based in part on strong benchmarking performance, there is the risk of a ``winner's curse'' \cite{zhong2010correcting} in which their accuracy may be artificially high in years that were included in the benchmarking work and the blended model may similarly overweight them.  Although the 1965-1978 hold-out period lies outside the AIWP models' training data, benchmarking during these years contributed to the selection of AIFS and NGCM. This bias is likely small, as the wider set of models examined during benchmarking generally exhibited similar levels of skill.

Finally, while the cross-validation procedure accounts for overfitting in a particular model specification, the selection of the best-performing model specification is still subject to a winner's curse and measures of its skill may be biased upwards. \cite{varma2006_cv_bias} This bias is generally small when comparing a limited number of candidate models and does not affect performance estimates on data not used for model selection.

\subsection*{Evaluation Metrics}
Models were evaluated across three standard probabilistic metrics: the Ranked Probability Score (RPS), the Brier Score, and the Area Under the Receiver Operating Characteristic Curve (AUC). Overall results are provided separately for the 2000-2024 model selection period, the 1965-1978 hold-out period, and the 2025 dissemination period.

As above, for each forecast $i$ and bin $j$ let $p_{ij}$ denote the forecast probability of onset in this bin, and set $Y_{ij} = 1$ if onset did occur in this bin, and 0 if not. The following metrics were used:
\begin{itemize}
    \item \textbf{Brier score (BS)}: 
\begin{equation}
    \text{BS} = \dfrac{1}{n} \sum_{i = 1}^n \sum_{j = 1}^m \left(Y_{ij} - p_{ij} \right)^2
\end{equation}

where $n$ is the number of forecasts and $m$ is the number of bins per forecast.   
\item \textbf{Ranked Probability Score (RPS):} The RPS is a generalization of the Brier Score which takes into account the distance between bins: for example, a prediction of onset in week 2 is ``closer'' to correctly predicting a week 3 onset than a prediction of onset in week 1. It is defined as 
\begin{equation}
\text{RPS} = \dfrac{1}{n} \sum_{i = 1}^n \sum_{k = 1}^m \left(\sum_{j = 1}^k \left(Y_{ij} - p_{ij}\right) \right)^2    
\end{equation}
 \item \textbf{Area Under the Receiver Operating Characteristic Curve (AUC): } The Receiver Operating Characteristic (ROC) curve is formed by using a probability threshold $t\in[0,1]$ for a binary classifier based on the model output and plotting the true positive rate against the false positive rate for all choices of $t$. These points define a curve, called the ROC curve, and the area under the curve is bounded by 0 and 1. For constructing this curve, we treat each bin's forecast as a separate binary forecast.   The area under this curve can be interpreted as the probability that, selecting a random forecast and bin in which onset occurred and selecting a random forecast and bin in which onset did not occur, the model assigned higher probability to the bin in which onset occur than the bin in which onset did not occur. More formally, this can be computed as
 \begin{equation}
 \text{AUC}  = \dfrac{\displaystyle\sum_{i,j,i',j'} Y_{ij}(1-Y_{i'j'}) \cdot 1[p_{ij} > p_{i'j'}]}{\displaystyle\left(\sum_{i,j} Y_{ij}\right)\left(\sum_{i,j} (1- Y_{ij})\right)}.    
 \end{equation}

 In general, improving the resolution of a forecast will increase its AUC, but improving its calibration without improving its resolution will not. Note that a higher AUC corresponds to a better forecast. 
\end{itemize}

Brier scores and ranked probability scores are then converted to \emph{skill} scores (BSS and RPSS), defined as the percent improvement a given metric shows relative to unconditional climatology. Formally, for a metric $x$ and set of forecasts $I$, we define
\begin{equation}
\text{Skill}(x,I) = 1 - \dfrac{x_I}{x_{\text{climatology, I}}}  
\end{equation}
 We did not adjust for ensemble size, as standard adjustments do not apply to calibrated outputs and so would not affect any of our main results. 

\subsection*{Model Evaluation}
Model skill was calculated according to the three metrics above during the 2000-2024 model selection period, a 1965-1978 hold-out period, and finally during the 2025 dissemination period. All skill scores are computed relative to static climatology. Note that the decision-theory criteria for dissemination are then that a model's skill scores should exceed those of the evolving expectations model, \emph{not} that they should exceed zero. 

After model selection was complete, the blended model was compared to raw NGCM model outputs, raw NGCM model outputs subject to a simple calibration scheme, or a traditional multimodel ensemble on the fuller 2000-2024 period. Calibration was performed via Platt Scaling\cite{platt1999probabilistic} on individual bins and renormalizing so that probabilities add up to one, and cross-validated by year. Since the AIWP models do not explicitly forecast the MOK component of our onset definition, we evaluated these outputs using a climatological median MOK date of June 2. Sensitivity analyses were conducted to check that results were not driven by this component of the definition. When a model output identified a potential onset without a full 30-day followup period, we counted this as a forecasted onset unless the model forecasted a 10-day dry spell before the end of the forecasting window.  

To make the comparison to a multi-model ensemble as favorable as possible for the ensemble, we identified the post-hoc ensemble weights that maximized the RPSS in the 2000-2024 period, using grid search on weights in 10 percentage point increments to find coarsely estimated optima, and then using a standard Broyden–Fletcher–Goldfarb–Shanno (BFGS) algorithm \cite{broyden1970convergence, fletcher1970new, goldfarb1970family, shanno1970conditioning} initialized at these weights to refine these estimates into the best post-hoc weights overall. 

To understand whether model skill was likely due to overfitting, we also evaluated the skill of various submodels (excluding the 10-day predictors or only including a subset of five-day predictors) on both the model selection period and the additional hold-out period. (Fig. \ref{fig:fig4}, Extended Data Fig. 4) 

\subsection*{Evaluation of 2025 Dissemination}
Only 28 grid cells actually received forecasts, because the monsoon had progressed past the southernmost grid cells before dissemination began. Following discussions with stakeholders, dissemination in each grid cell continued until the IMD had declared onset over at least half of the cell. Since the IMD's meteorology-based declaration and the purely rainfall-based agricultural definition of onset can differ, this led to situations in which forecasts went out after the Moron-Robertson onset had already triggered but before the meteorological onset had been declared. In seven grid cells the Moron-Robertson onset occurred before any forecasts went out, while five additional grid cells had a single forecast go out 1-3 days after the Moron-Robertson onset. These forecasts are excluded from the evaluation as they are not directly related to the performance of the forecasting model, but highlight operational challenges inherent to large-scale dissemination. 

\subsection*{Decision-Theory Model}
In this section we adapt standard tools in decision theory\cite{blackwell1953equivalent,epstein1962_bayesian_meteorology, murphy1977_value_of_forecasts} to apply to the problem of a weather forecaster choosing what information to provide in order to benefit a heterogeneous group of farmers. The model laid out below is designed to capture a setting in which:
\begin{enumerate}
    \item The forecaster has information about future weather that farmers do not have. Farmers have to make decisions for which the optimal choice depends both on the weather and on information about their own circumstances the forecaster does not have. The model also applies in cases where the forecaster has access to private information, but cannot perfectly tailor messages to individual farmers' circumstances due to e.g. limitations of the dissemination platform. 
    \item Communication is restricted to (potentially probabilistic) forecasts: forecasters cannot fully explain the forecasting process or provide individualized advice, but may convey uncertainty through probabilities.
    \item Farmers have some information about the weather (for example, what happened in past years). Some farmers believe the forecaster has already incorporated this information into the forecasts and therefore they take the forecasted probabilities at face value. In other words, these farmers adopt the forecasted probabilities as their own beliefs without further modification.  
\end{enumerate}
The model allows for risk aversion and does not place restrictions on the value of information, actions, or payoffs. All of our results hold for farmers who have information about the weather which they correctly incorporate into their posterior beliefs via Bayesian updates; we focus on farmers who take the forecasts at face value to illustrate the stronger assumptions needed for forecasts to provide value to these users. 

Formally, assume farmer $i$ from a set $\mathcal{I}$ chooses an action $\theta_i \in \Theta_i$, whose expected payoff $g_i(\theta_i, \omega_i)$ depends on future weather state $\omega_i \in \Omega_i$. This payoff differs between farmers, since their circumstances differ, and each farmer tries to maximize their own payoff. As an example, $\theta_i$ may represent a crop or fertilizer choice or planting date, while $\omega_i$ may denote the onset date of sustained rainfall or the expected amount of rainfall next week in farmer $i$’s location. We assume farmers' decisions affect their payoffs but not those of other farmers. 

Some information, given by the value of a random variable $S_i$, is known to both farmer $i$ and the forecaster. The forecaster has access to a additional weather information in the form of a stochastic forecast scheme $\Xi_i$. A realization of $\Xi_i$ is a probabilistic forecast $\xi_i : \Omega_i \to [0,1]$.  For tractability, we assume that $\Omega_i$ is finite and that the forecaster must commit ex ante to whether a forecast is sent, prior to observing the realization $\xi_i$. Farmers have private information about agriculture and their own circumstances, so the payoffs $g_i$ are known to individual farmers but not the forecaster.

The forecaster is benevolent in the sense that they want to maximize farmers' payoffs. A forecast is defined to be \emph{useful} if no farmer is worse off (in expectation) for having received forecasts, and at least one farmer is better off in the sense of being able to make a decision with higher expected payoff.

Farmer $i$'s prior belief $p_i$ over $\Omega_i$ is given by
\[
p_i(t) = \mathbb{P}(\omega_i = t \mid  S_i = \sigma_i).
\]

We say that the forecast scheme $\Xi_i$ is \emph{well-calibrated} if the probabilities produced by its forecasts match the actual weather variable's empirical frequencies; that is, for any potential forecast $\xi_i$,
\[
\mathbb{P}(\omega_i = t \mid \Xi_i = \xi_i) = \xi_i(t)
\quad \forall t \in \Omega_i .
\]

We say that the forecast scheme \emph{incorporates all information farmer $i$ believes to be shared} if conditioning on the information farmer $i$ believes to be shared does not give any information beyond the forecasted probabilities; in other words, if for all possible forecasts $\xi_i$ and shared information $\sigma_i$
\[
\mathbb{P}(\omega_i = t \mid \Xi_i = \xi_i, S_i = \sigma_i)
=
\mathbb{P}(\omega_i = t \mid \Xi_i = \xi_i).
\]

On the other hand, the forecast scheme \emph{contains information not previously known to the farmer} if there is a piece of information the the farmer believes to be shared, and a forecast in the forecasting scheme, such that the probability of some weather event is changed by knowing the piece of shared information. In other words, if there exist $\xi_i$, $\sigma_i$, and $\pi_i$ such that
\[
\mathbb{P}(\omega_i = t \mid \Xi_i = \xi_i, S_i = \sigma_i)
\neq
\mathbb{P}(\omega_i = t \mid S_i = \sigma_i).
\]

The farmer adopts the forecast as their posterior $\tilde{p}(t) = \xi(t).$

\subsection*{Theoretical Results}
In this section we state the main results of the decision-theory model. Proofs can be found in Supplementary Note 1. We first show that each farmer is weakly better off if the forecaster provides probabilistic forecasts rather than deterministic forecasts. This is a straightforward application of Blackwell's Theorem\cite{blackwell1953equivalent}. Intuitively, different farmers may have different thresholds for action (depending on payoffs, inputs, access to outside options, etc.) and so it may not be possible to find a single deterministic threshold which is optimal for all farmers. Providing probabilities allows individual farmers to make use of the information in different ways depending on their circumstances. 
\begin{prop}\label{probabilistic}
Suppose forecasts are well-calibrated. Then each farmer is (weakly) better in expectation to if the forecaster provides them with the probabilities of each potential outcome rather than coarsening them to deterministic forecasts. 
\end{prop}

We then examine when farmers benefit from receiving forecasts. As is common in decision-theory literature, we can give very general conditions under which farmers \emph{weakly} benefit (i.e. are not harmed), but will need to assume the set of decisions is large and diverse to make general statements about \emph{strict} benefits. This is plausible in a complex and heterogeneous setting like agriculture and is empirically testable (Proposition \ref{sufficient}.)

\begin{prop}\label{benefit}
If the forecasting scheme is well-calibrated and incorporates all information each of the farmers believes to be shared, then:
\begin{enumerate}[label=\alph*.]
    \item All farmers weakly benefit in expectation from receiving forecasts
    \item Individual farmers can only strictly benefit if the forecast scheme contains information not previously known to them. 
     \item For each farmer $i$ for whom the forecast scheme contains previously unknown information, there exists a payoff function $g_i(\theta_i,\omega_i)$ such that they would strictly benefit in expectation if their payoff function were $g_i.$
\end{enumerate} 
\end{prop}
Informally, we interpret the third statement as saying at least one farmer strictly benefits from a forecast containing new information as long as the set of decision problems (objectives, inputs, outside options, risk thresholds, etc.) is sufficiently diverse. 

Finally, we give some results related to the question of measurement. We first show that, even if policymakers could measure farmer payoffs perfectly, the realized value of the forecasts depends on the weather realization, so an individual randomized controlled trial will not allow an analyst to estimate the overall expected effects of providing forecasts.  We then show that under the assumptions we have made so far, and an additional assumption that decisions are impactful, the \emph{number} of farmers who have changed a decision in response to forecasts provides a lower bound for the number of farmers who benefit from the program. Finally, we show that if the forecaster can identify all decisions impacted by the forecast, measuring changes in decision-making has higher statistical power for detecting whether forecasts are useful than measuring changes in outcomes. 

For simplicity, we assume a finite number of farmers $i = 1,2,\cdots, n,$ although similar statements hold for more general sets. Throughout what follows we assume forecasts are well-calibrated and fully incorporate shared information. 

\begin{prop}\label{no-outcome-rct}
Suppose the forecaster has access to a measure of farmer payoffs $g$, and creates an experiment by which farmers are randomly assigned to receive or not receive forecasts. The confidence interval produced by this experiment will in general not be a confidence interval for the average expected utility gain from forecasts at any confidence level. 
\end{prop}

To illustrate, suppose the forecast concerns a large-scale natural disaster (e.g., a regional flood) that affects all farmers in the experiment simultaneously. In some years, the forecasted probability of a flood may be close to the baseline probability, in which case the forecasts might have limited value to farmers who already had this information. In other years, the forecast may assign a high probability (say, 80 percent) to a flood. The expected value of such information depends on how farmers trade off the costs of precaution against the expected losses from flooding, and hence on the full distribution of outcomes induced by the forecast. However, in any realized experiment only a single aggregate outcome is observed: either the flood occurs or it does not. Consider the case where it does not. Even with an arbitrarily large sample of farmers in a single season the experiment compares the costs of precautions farmers took given an $80\%$ risk of flooding with the costs incurred due to the actual non-occurrence of a flood. These will not match the expected payoff to farmers of receiving forecasts unless the experiment is repeated over a number of years and geographic regions and results are pooled across different realized outcomes\cite{rosenzweig2020external}. 

Risk-averse farmers may be modeled as having a payoff function $g_i(\theta_i, \omega_i) = u(h_i(\theta_i, \omega_i)),$  where $h$ is an observable function (e.g. income) and $u$ is a concave utility function. In this case, measurements of the impact of forecasts on $h$ may be negative even when the effect on farmers is positive. Intuitively, this is because farmers may use forecasts to manage risk, paying a small cost to insure themselves against large negative shocks. 
\begin{prop}\label{manage-risk}
If farmers are risk-averse in the sense of the preceding paragraph, then a forecast can make them better off in the sense of increasing their expected utility, reducing their expected observed income.
\end{prop}

Rather than measuring payoffs directly, we find that evaluators can learn about the value of forecasts by measuring whether or not forecasts change farmers' decisions
\begin{prop}\label{sufficient}
If forecasts are well-calibrated and fully incorporate all information farmers believe to be shared, and that the utility-maximizing decision conditional on either the prior or the received forecast is unique with probability one.  Then, the number of farmers who strictly benefit in expectation from a forecasting scheme is greater than or equal to the expected number (across all draws from the scheme) of farmers who changed a decision as a result of a single instantiation of the forecast.
\end{prop}

The new assumption added in this proposition, that utility-maximizing decisions are unique with probability one, is primarily a technical assumption to remove cases where forecasts spuriously change a decision with no impact on payoff. Finally, we show that measuring decisions can improve power over measuring yield or income outcomes.

\begin{prop}\label{decision-power}
Suppose a forecaster can perfectly measure farmer choices $\theta_i$ and payoffs $g_i$ and runs a randomized controlled trial to determine whether any farmers \emph{strictly} benefit from forecasts. Under the assumptions of Proposition \ref{sufficient}, it follows that:
\begin{enumerate}
    \item If the distribution of either $\theta_i$ or $g_i$ is changed by the forecasts, then at least one farmer strictly benefits.
    \item For any test of whether the distribution of $g_i$ is changed by the forecasts based on the experiment, there is a test of whether the distribution of $\theta_i$ is changed by the forecasts with weakly lower Type I and Type II error rates in expectation.
\end{enumerate}
\end{prop}

\section*{Acknowledgments}
This work grew out of a collaboration with India's Ministry of Agriculture and Farmers' Welfare. We thank Joy Dada and Stephanie Dragoi for excellent research assistance. We are grateful for fruitful discussions with Shreya Agrawal, Medha Deshpande, Josh Deutschmann, Genevieve Flaspohler, Subimal Ghosh, Ankur Gupta,  Neil Hausmann, Stephan Hoyer, Peter Huybers, Mrutyunjay Mohapatra, Vincent Moron, Raghu Murtugudde, Yaw Nyarko, Sivananda Pai, D.R. Pattanaik, Thara Prabhakaran, Satya Prakash,  V.S. Prasad, Suryachandra Rao, Muthalagu Ravichandran, Sakha Sanap, Sahadat Sarkar, K. K. Singh, Gokul Tamilselvam, Witold Więcek,  and Janni Yuval. We are grateful to Google and the European Centre for Medium-Range Weather Forecasts for making the AI models used in this work publicly available, to Shreya Agrawal and Stephan Hoyer for assistance in running NGCM with real-time initial conditions, and to the India Meteorological Department for publishing 125 years of rain gauge data. We thank the many people at the Development Innovation Lab, the Development Innovation Lab - India, and Precision Development for their work preparing forecasts for dissemination. This work is partially supported by AIM for Scale, the Gates Foundation, Wellspring Philanthropic Fund, and by the University of Chicago's Human-centered Weather Forecasts Initiative (a program of the Institute for Climate and Sustainable Growth, ICSG), Development Innovation Lab, and AI for Climate Initiative (a joint program of the Data Science Institute, DSI, and ICSG). We thank the University of Chicago's DSI, Research Computing Center, and Social Science Computing Services for computational resources.

\section*{Author contributions}
PH, WB, AJ, and MK conceived and planned the study and supervised the research. AM generated the AIWP forecasts. CA and MK conceptualized the decision-theory framework, and CA proved the associated theorems. CA, RM, AM, MG, and TY analyzed data. CA designed the ``evolving expectations'' and blended models. CA, PH, WB, AJ, and MK drafted the manuscript. CA and AJ created the figures. All authors discussed and interpreted the results. All authors reviewed and edited the manuscript.

\section*{Data availability}
ERA5 data used in this study for initializing the AI models is obtained from the Copernicus Climate Change Service (C3S) Climate Data Store. Real-time initial conditions used in the 2025 monsoon season forecasts can be sourced from NCEP GDAS/FNL product inventory (\url{https://www.nco.ncep.noaa.gov/pmb/products/gfs/#GDAS}) and the ECMWF IFS Open Data Portal (\url{https://data.ecmwf.int/forecasts/}). The India Meteorological Department (IMD) 1-degree gridded rainfall data are available from the IMD data portal (\url{https://www.imdpune.gov.in/cmpg/Griddata/Rainfall_1_NetCDF.html}). ECMWF's AIFS model weights are obtained from \url{https://huggingface.co/ecmwf/aifs-single-1.0}. Pre-trained weights for Google Research's NGCM (stochastic, IMERG precipitation) can be accessed at \url{https://neuralgcm.readthedocs.io/en/latest/checkpoints.html}. The re-gridded forecasts from NGCM and AIFS and the code used in the study can be accessed at \url{https://doi.org/10.5281/zenodo.18894299}. During the 2025 season, blended model outputs were uploaded to Zenodo as they were produced; see \url{https://zenodo.org/records/15477259}, \url{https://zenodo.org/records/15490559}, \url{https://zenodo.org/records/15527948}, \url{https://zenodo.org/records/15588265} \url{https://zenodo.org/records/15632841}. \url{https://zenodo.org/records/15651248},  \url{https://zenodo.org/records/15683783},  \url{https://zenodo.org/records/15731043}, and \url{https://zenodo.org/records/15756597}. Not all forecasts uploaded to Zenodo were disseminated. Blended model outputs will also be placed on Zenodo during the 2026 season. 
\section*{Figures}

\newpage 
\clearpage

\captionsetup[table]{name=Extended Data Table}
\captionsetup[figure]{name=Extended Data Fig.}
\setcounter{figure}{0}

\begin{figure}[htbp]
  \centering
  \includegraphics[width=\textwidth]{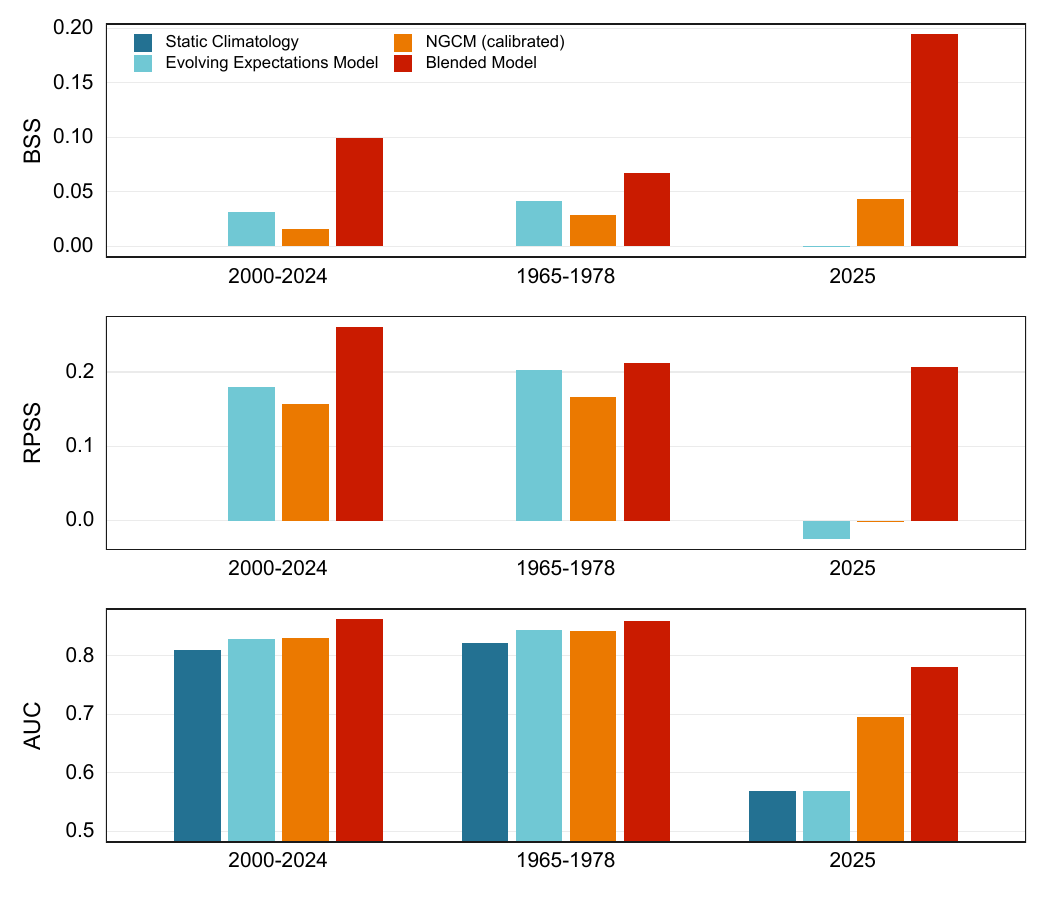}
\caption{\textbf{Skill scores aggregated by test period.} The blended model is cross-validated during the 2000-2024 period, and trained on 2000-2024 data during the other periods. The climatology model and the evolving expectations model are cross-validated by year using 1900-2024 IMD data. Skill scores are all computed relative to static climatology.} 
\end{figure}



\begin{figure}[htbp]
  \centering
  \includegraphics[width=\textwidth]{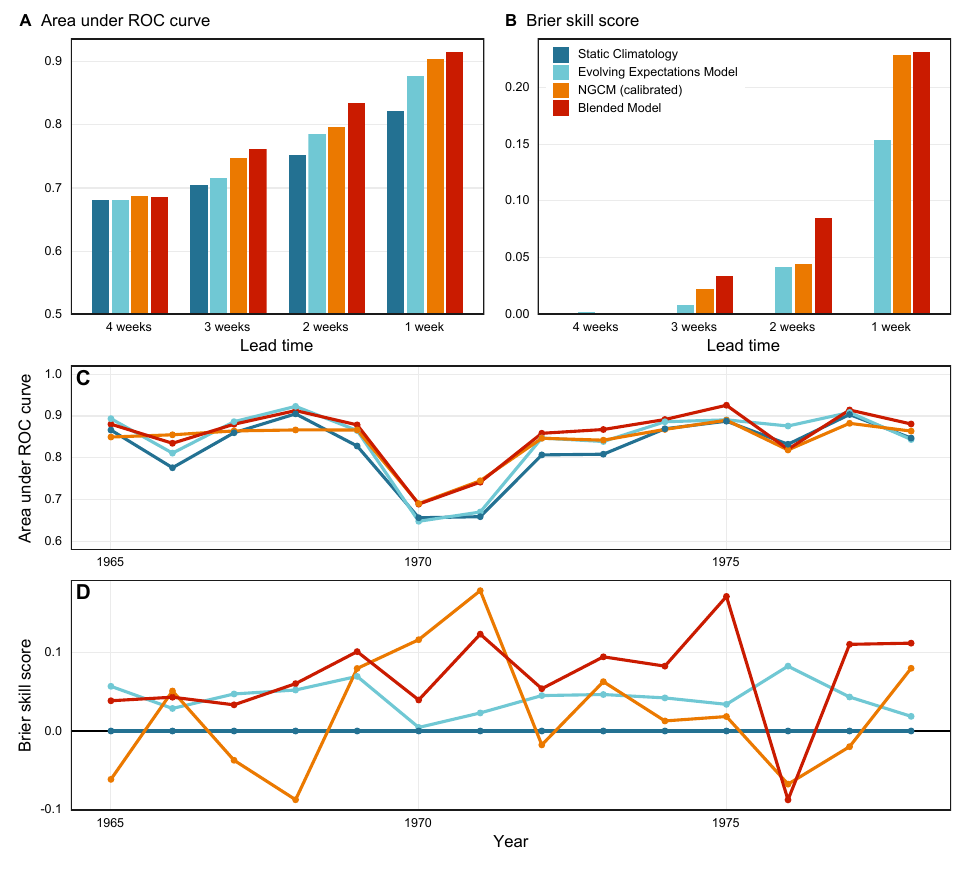}
\caption{\textbf{Temporal components of model skill, 1965-1978.} \textbf{A)} Area under ROC curve (AUC) by lead time during the 1965-1978 pre-satellite-era period. The baseline for the bars is chosen to be 0.5, indicating the AUC of a forecast with no ability to distinguish onsets from non-onsets. \textbf{B)} Brier skill score by lead time, computed relative to a traditional (static) climatology model. \textbf{C)} Area under ROC curve by year across all lead times. The 2000-2024 scores are computed via cross-validation. \textbf{D)} Brier skill score by year across all lead times, calculated relative to a traditional (static) climatology model. } 
\end{figure}

\begin{figure}[htbp]
  \centering
  \includegraphics[width=\textwidth]{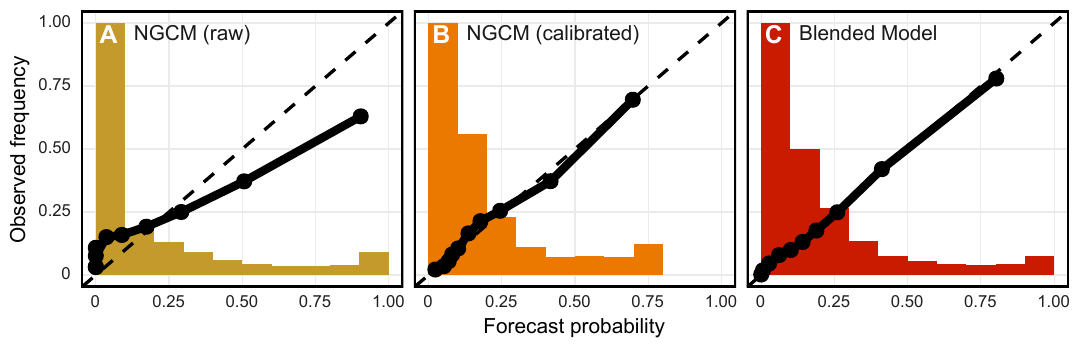}
  \caption{\textbf{Reliability diagram and histogram of probabilities assigned by different models during the 1965-1978 pre-satellite-era period.} If the model is well-calibrated the points should line up with the dashed line.  Each point represents a decile of probabilities assigned by the model, and compares the average predicted probability in that decile and the fraction of events which occurred in observation. Histograms are normalized so that the bin capturing probabilities between 0 and 10\% has height equal to one.}
\end{figure}

\begin{figure}[htbp]
  \centering
  \includegraphics[width=\textwidth]{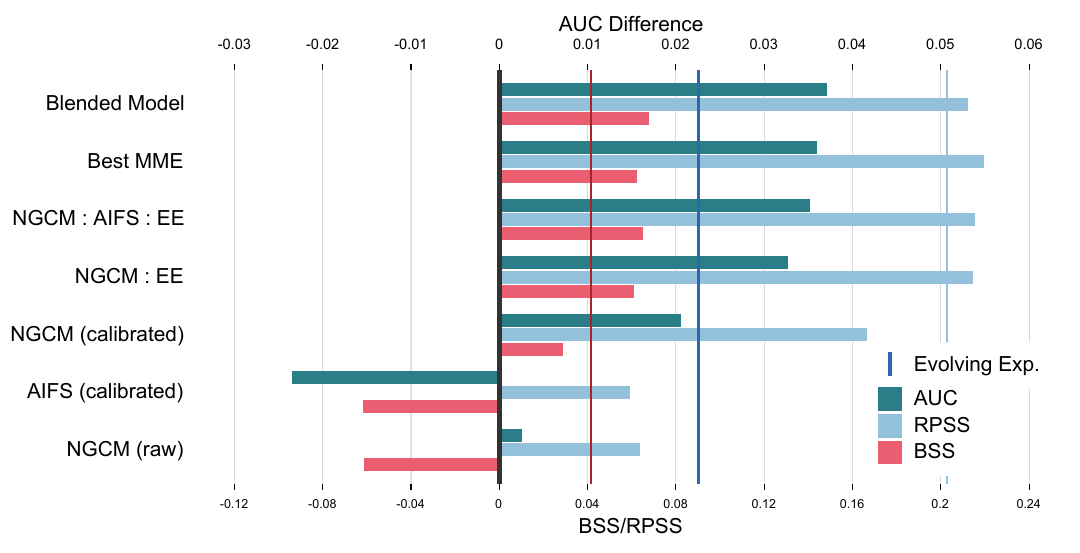}
  \caption{\textbf{Performance of models during the 1965-1978 pre-satellite-era period, including submodels of the blended model.} Vertical lines represent the evolving-expectations (EE) model, which our decision-theory model implies should be a baseline for dissemination. The interaction models, indicated with colons, are submodels of the final blended model using 5-day rainfall variables from one or more AIWP models but not 10-day rainfall variables. The best multimodel ensemble is selected as a linear combination of probabilities from AIFS, NGCM, and the evolving expectations model maximizing the RPSS post-hoc on the 2000-2024 period. All other models (including calibration) are trained via leave-one-year-out cross-validation. Skill scores and differences in AUC are computed relative to unconditional climatology, which has an AUC of 0.822, a Brier Score of 0.567, and a Ranked Probability Score of 0.594. }
\end{figure}

\begin{figure}[htbp]
  \centering
  \includegraphics[width=\textwidth]{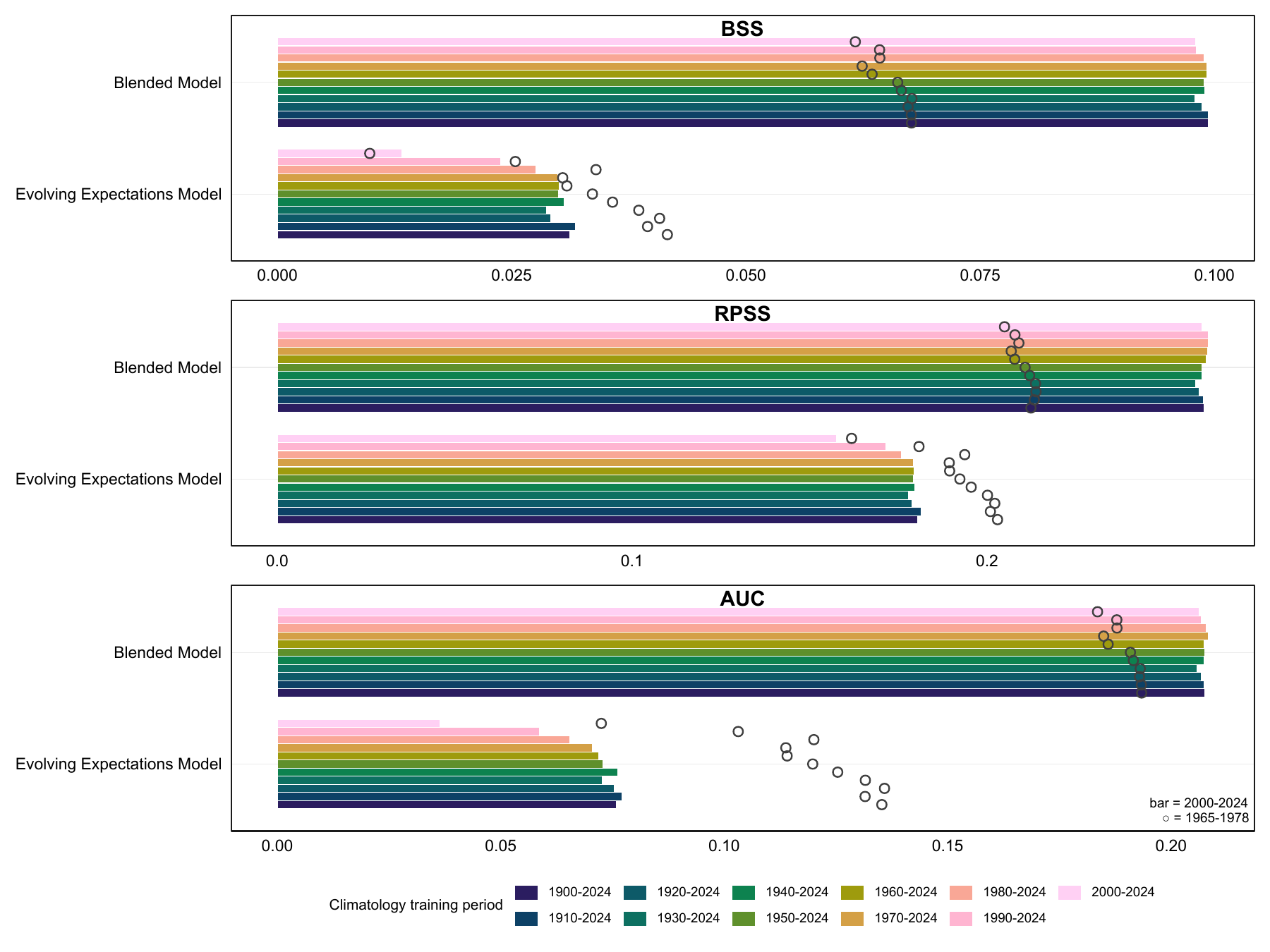}
  \caption{Skills of the evolving expectations model and the blended model using different training periods for the evolving expectations model. Skill scores are computed relative to static climatology trained on all years 1900-2024. Bars indicate scores on the 2000-2024 period, and circles represent scores in the 1965-1978 period. }

\end{figure}

\newpage 
\bibliography{blend_refs, AIWP}


\end{document}


\maketitle
\thispagestyle{empty}

\noindent\textsuperscript{1}Development Innovation Lab, University of Chicago, IL, 60637\\
\textsuperscript{2}Department of the Geophysical Sciences, University of Chicago, IL, 60637\\
\textsuperscript{3}Data Science Institute, University of Chicago, IL, 60637\\
\textsuperscript{4}Development Innovation Lab India, University of Chicago Trust, India, 560025\\
\textsuperscript{5}Department of Earth and Planetary Science, University of California, Berkeley, California, 94720\\
\textsuperscript{6}Kenneth C. Griffin Department of Economics, University of Chicago, IL, 60637\\
\textsuperscript{7}Harris School of Public Policy, University of Chicago, IL, 60637\\
\textsuperscript{8}Climate and Ecosystem Sciences Division, Lawrence Berkeley National Laboratory, Berkeley, California, 94720\\

\newpage


\begin{itemize}
    \item Supplementary Notes S1 to S2
    \item Supplementary Figure S1 to S17
\end{itemize}

\clearpage

\section*{Supplementary Note 1: Proofs of Theoretical Results}
This section records proofs of the propositions in the "Theoretical Results" part of the method section. We first establish a Lemma that will underlie the strategy of most of the proofs. 
\begin{lem}\label{bayes}
If $\xi(t)$ is well-calibrated and fully incorporates all information farmer $i$ believes to be shared, then adopting the forecast at face value represents an accurate Bayesian update for the farmer. 
\end{lem}
\begin{proof}[Proof of Lemma \ref{bayes}]
The given assumptions imply
\begin{align*}
\mathbb{P}(\omega_i = t \mid \Xi_i = \xi_i, S_i = \sigma_i) &= \mathbb{P}(\omega_i = t \mid \Xi_i = \xi_i) \\
&= \xi_i(t).
\end{align*} 
\end{proof}

\begin{proof}[Proof of Proposition 1]
This is essentially one direction of Blackwell's informativeness theorem \cite{blackwell1953equivalent}. Let $\hat{\xi}$ denote a deterministic forecast computed from a forecast instance $\xi$. The definition of $\theta_{i,f}$ implies that 
\[
\EE_{\omega \sim \xi} [g_i(\theta_{i,\xi}, \omega)] \geq \EE_{\omega \sim \xi} [g_i(\theta_{i,\hat{\xi}}, \omega)]
\]
In particular, 
\[
\EE_\xi[ \EE_{\omega \sim \xi} [g_i(\theta_{i,\xi}, \omega)] \mid \xi] \geq \EE_\xi[ \EE_{\omega \sim \xi} [g_i(\theta_{i,\hat{\xi}}, \omega)] \mid \xi]
\] 
so the result follows from the Law of Iterated Expectations. 

\end{proof}

The proof of Proposition 2 consists of Lemma \ref{bayes} applied to  Lemma \ref{bayes-benefit}.

\begin{lem}\label{bayes-benefit}
The statement of Proposition 2 holds for farmers who perform perfect Bayesian updates without the assumption that forecasts are well-calibrated or incorporate shared information.
\end{lem}
\begin{proof}
For any distribution $f$ on $\Omega_i$, let $\theta_{i,f}$ be farmer $i$'s decision maximizing expected utility $$\EE_{\omega_i \sim f}[g_i(\theta_{i,f}, \omega_i)]$$ under the assumption that $\omega_i$ is distributed according to $f$.  The expected benefit of a forecast scheme $\Xi_i$ to farmer $i$ is given by
\[
\EE[ g_i(\theta_{i,\tilde{p}_i}, \omega_i) - g_i(\theta_{i,p_i},\omega_i)],
\]

which can be rewritten as
\[
\EE_{\Xi_i , S_i} \left[ \EE_{\omega_i \sim \tilde{p}_i \mid \Xi_i, S_i}[ g_i(\theta_{i,\tilde{p}_i}, \omega_i) - g_i(\theta_{i,p_i},\omega_i) ]\right],
\]
This is guaranteed to be non-negative by definition $\theta_{i, \tilde{p}_i}$ is the choice of $\theta$ maximizing expected utility when $\omega_i$ is distributed according to $\tilde{p}_i.$  If the forecast does not contain information outside the farmer's decision set, then $\tilde{p}_i(t) = p_i(t)$ for all $i$ and $t$, and no farmer will strictly benefit. By Blackwell's informativeness theorem \cite{blackwell1953equivalent}, if the forecasts contain information outside the prior information set at least one decision problem will be improved by the forecast, so the converse follows as long as the set of decision problems is sufficiently broad to contain the one guaranteed by Blackwell's Theorem. 
\end{proof}

\begin{proof}[Proof of Proposition 3]
To provide a counterexample, we assume a large number of identical farmers who receive the same $\xi_i$ and $\omega_i,$ and whose $g_i$'s are identical but whose measured $\hat{g}_i$ may differ due to noise. (This latter assumption is not necessary but prevents the confidence interval from having size zero.)

The average expected utility gain from forecasts is
\[
 \EE_{\Xi_i , S_i} \left[ \EE_{\omega_i \sim \tilde{p}_i \mid \Xi_i, S_i}[ g_i(\theta_{i,\tilde{p}_i}, \omega_i) - g_i(\theta_{i,p_i},\omega_i) ]\right],
\]
which by assumption does not depend on $i$. Given a particular draw of $\xi_i$ and $\omega_i$, the measured treatment effect will be normally distributed with mean
\[
 g_i(\theta_{i,\tilde{p}_i}, \omega_i) - g_i(\theta_{i,p_i},\omega_i).
\]
and variance proportional to $1/\sqrt{n}.$ As $n \to \infty$, the confidence interval around this quantity will converge to length zero, so if the expected utility gain has any dependence on $\xi_i$ or $\omega_i$ it will exclude the true average expected utility gain. 
\end{proof}

\begin{proof}[Proof of Proposition 4]
It suffices to give an example. Suppose $u(x) = \sqrt{x}$, and suppose the farmer is deciding whether to purchase crop insurance against a potential drought.

Let $\omega_i$ be an indicator for whether or not a drought occurs, and let $\theta_i$ be an indicator for whether or not the farmer purchases insurance. We assume that a priori the probability of a drought is $10\%$, and the insurance is priced in a way that the farmer would not want to buy it without further information. Specifically, we set:
\begin{align*}
h(0,0) &= 100 \\
h(1,0) &= 0 \\
h(0,1) &= 81 \\
h(1,1) &= 16,
\end{align*}
so that the farmer's expected income without insurance is $90$ and with insurance is $74.5$, while their expected utility without insurance is $9$ and with insurance is $8.5.$

Now, suppose the forecasting system issues an accurate system about the likelihood of a drought. In particular, assume that $80\%$ of the time the system indicates a drought will not occur, and $20\%$ of the time the system indicates a drought will occur with probability $50\%$. In this case the farmers' expected income without insurance is the same, while their expected income if they buy insurance when the drought risk is elevated is 89.7. However, their expected utility if they buy insurance when the drought risk is elevated rises to 9.3.

In particular, a farmer with access to forecasts will manage their risk by buying the insurance plan, increasing their utility but decreasing their expected income. An experiment which only measures expected income would show that farmers were harmed, failing to account for their risk aversion. 
\end{proof}

\begin{proof}[Proof of Proposition 5]
Again we can assume farmers are Bayesian, appealing to Lemma \ref{bayes}. We saw above that the expected benefit of a forecast scheme $\Xi_i$ to such a farmer $i$ is 
\[
\EE_{\Xi_i , S_i} \left[ \EE_{\omega_i \sim \tilde{p}_i \mid \Xi_i, S_i}[ g_i(\theta_{i,\tilde{p}_i}, \omega_i) - g_i(\theta_{i,p_i},\omega_i) ]\right].
\]
Let $A_i$ be a random variable equal to $1$ if farmer $i$ changes a decision after seeing forecast $\xi_i$ and $0$ otherwise. The assumption on $\theta_{i, \tilde{p}_i}$ implies that for each $i$:
\[
\mathbb{P}(A_i = 1) = \mathbb{P}(g_i(\theta_{i,\tilde{p}_i}, \omega_i) > g_i(\theta_{i,p_i},\omega_i))
\]
Markov's inequality then implies that if $\mathbb{P}(A_i = 1) > 0$, then
\[
\EE_{\Xi_i , S_i} \left[ \EE_{\omega_i \sim \tilde{p}_i \mid \Xi_i, S_i}[ g_i(\theta_{i,\tilde{p}_i}, \omega_i) - g_i(\theta_{i,p_i},\omega_i) ]\right] > 0.
\]
so a farmer strictly benefits from the forecast scheme in expectation if their probability of changing a decision is positive.  The result then follows because
\begin{align*}
 \EE\left[\sum_i A_i \right]    &\leq \sum_i 1[E[A_i] > 0] 
\end{align*}
where the left hand denotes the expected number of farmers changing a decision from a single forecast instantiation and the right hand side is the number of farmers who benefit from the forecast scheme.
\end{proof}

\begin{proof}[Proof of Proposition 6]
The first part follows from Proposition 5, noting that $g_i$ can only be changed by the forecasts if $\theta_i$ is. The second follows from Blackwell's informativeness theorem, since $g_i$ is a garbling of $\theta_i$. The specific translation from informativeness to Type-I and Type-II errors can be found in \cite{blackwell1979theory} with further explanation in \cite{mu2021blackwell}.
\end{proof}

\section*{Supplementary Note 2: Sensitivity Analyses}
One potential concern is the gap in performance between NeuralGCM (and models built on it, including the multimodel ensemble) where ensemble member onsets are defined using a climatological MOK filter, and a variant where the true MOK is used (even for forecasts well before it occurs.)  To address this, we provide alternate versions of Figures 2 through 4 as well as Extended Data Figure 1 for alternate onset definitions, first by replacing the condition that onset must happen after MOK with a condition that it must happen after the climatological June 1 MOK date, and second by removing the MOK filter altogether and counting any date after May 1. 

The resulting figures (SI Figures S5 - S17) look generally similar to the figures in the main text. In particular, the observation that during the 2000-2024 period NeuralGCM does not outperform the evolving expectations model beyond one- or two- week lead times even after calibration, and the observation that the hybrid model outperforms any multimodel ensemble in the 2000-2024 period do not appear to be the result of an unfair comparison. A few comparisons, which are closer in the main text, appear to have some dependence on chosen definitions.

\clearpage

\section*{Supplementary Figures}

\begin{figure}[htbp]
  \centering
  \includegraphics[
    width=\textwidth,
    keepaspectratio
  ]{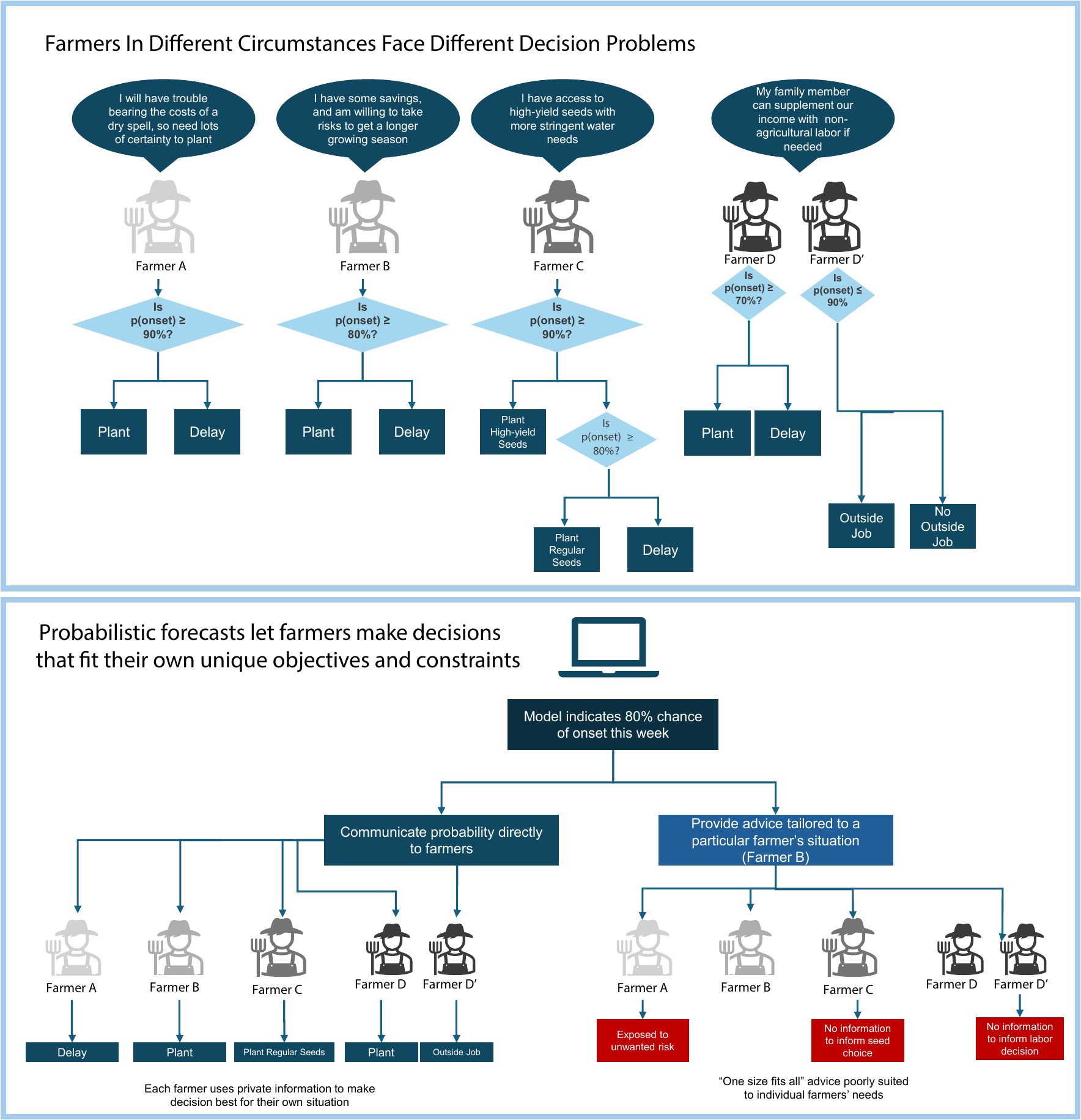}
  \caption{An illustration of the decision-theory model's implication that probabilistic forecasts provide more value than deterministic forecasts.}
\end{figure}

\setcounter{figure}{0}
\begin{figure}[htbp]
  \centering
  \includegraphics[width=\textwidth]{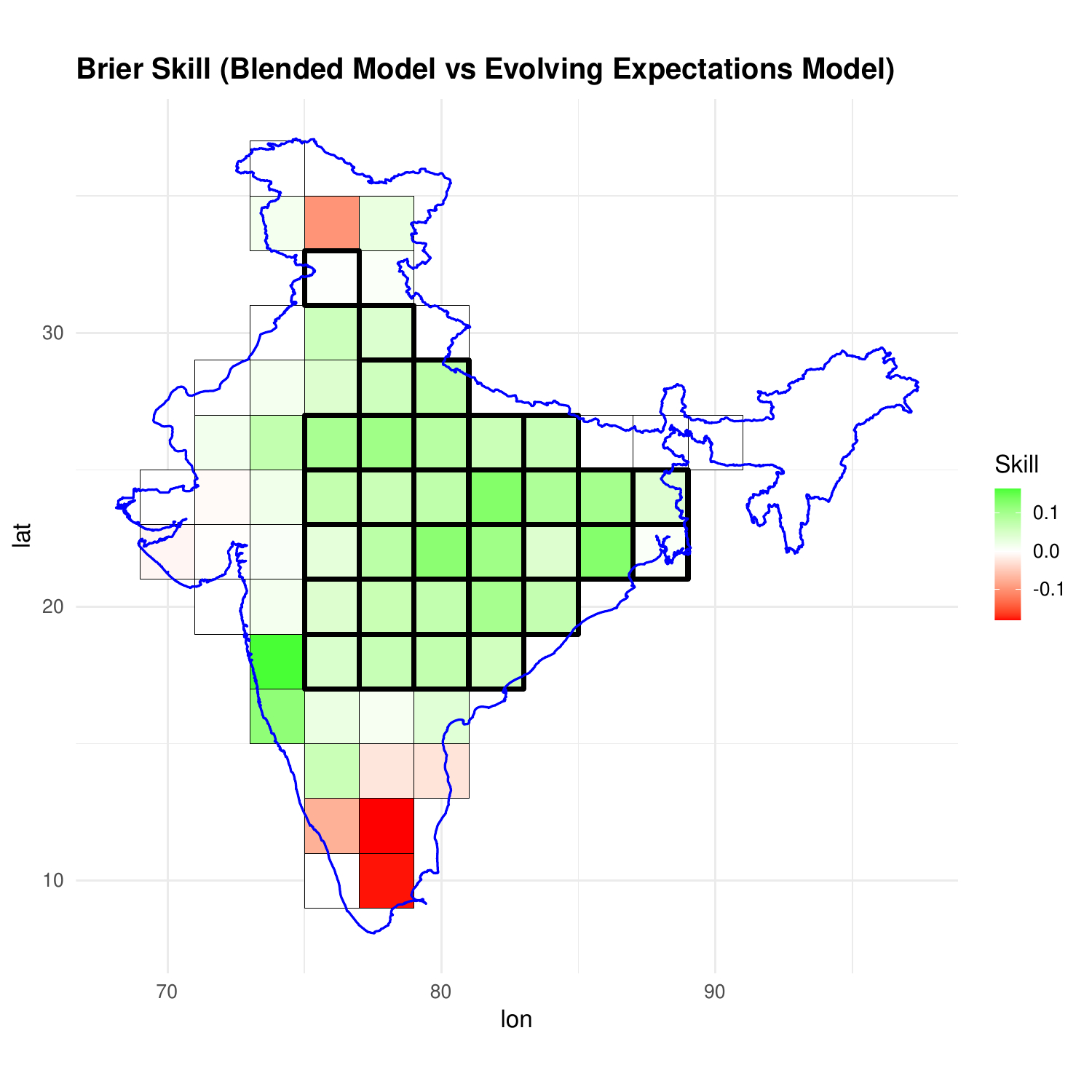}
  \caption{Map of Brier skill scores of blended model relative to the evolving expectations model by grid cell during the 2000-2024 cross-validation period. Grid cells used for model training and intended for dissemination are depicted with thick outlines. }
\end{figure}

\begin{figure}[htbp]
  \centering
  \includegraphics[width=\textwidth]{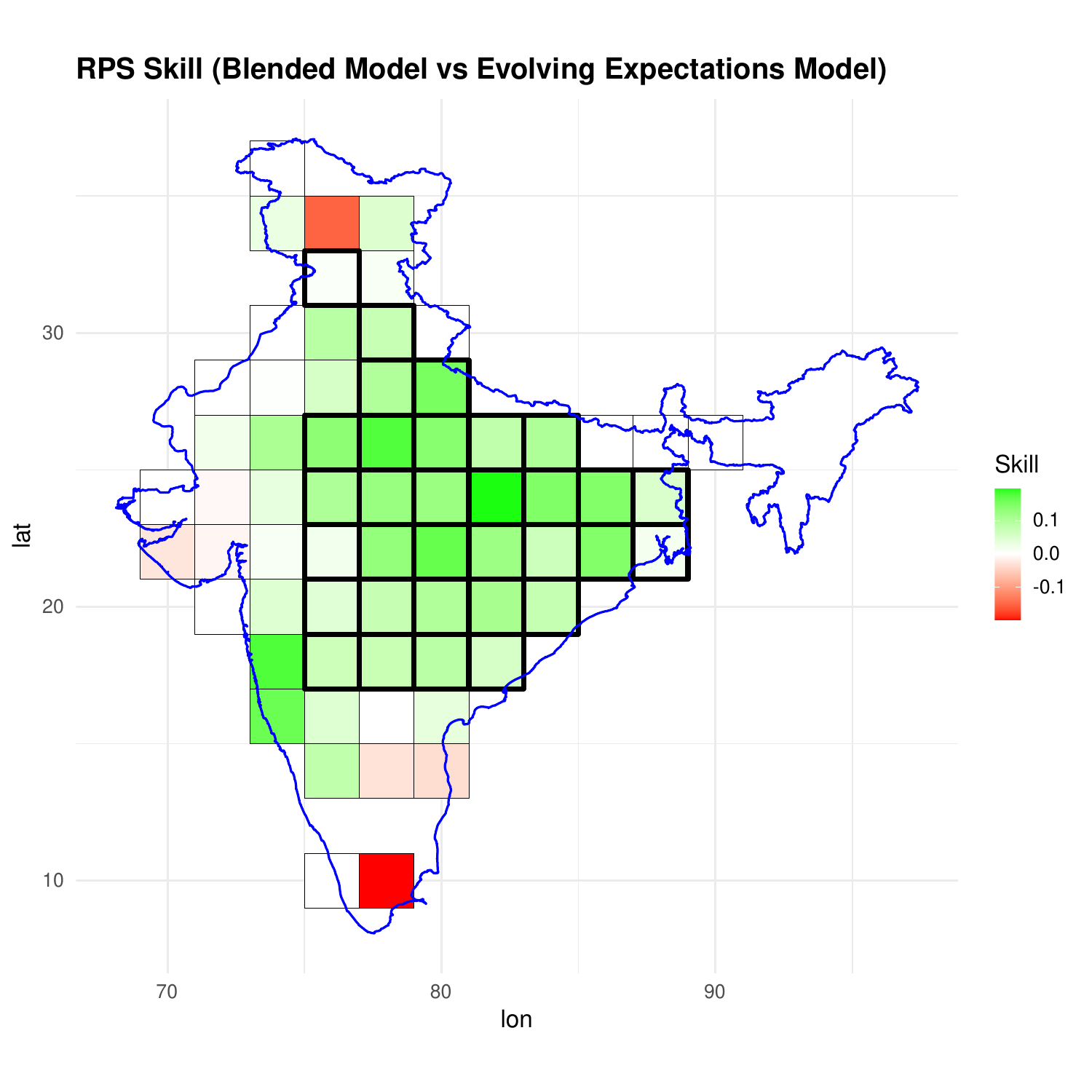}
  \caption{Map of RPS skill scores of blended model relative to the evolving expectations model by grid cell during the 2000-2024 cross-validation period. Grid cells used for model training and intended for dissemination are depicted with thick outlines. }
\end{figure}

\begin{figure}[htbp]
  \centering
  \includegraphics[width=\textwidth]{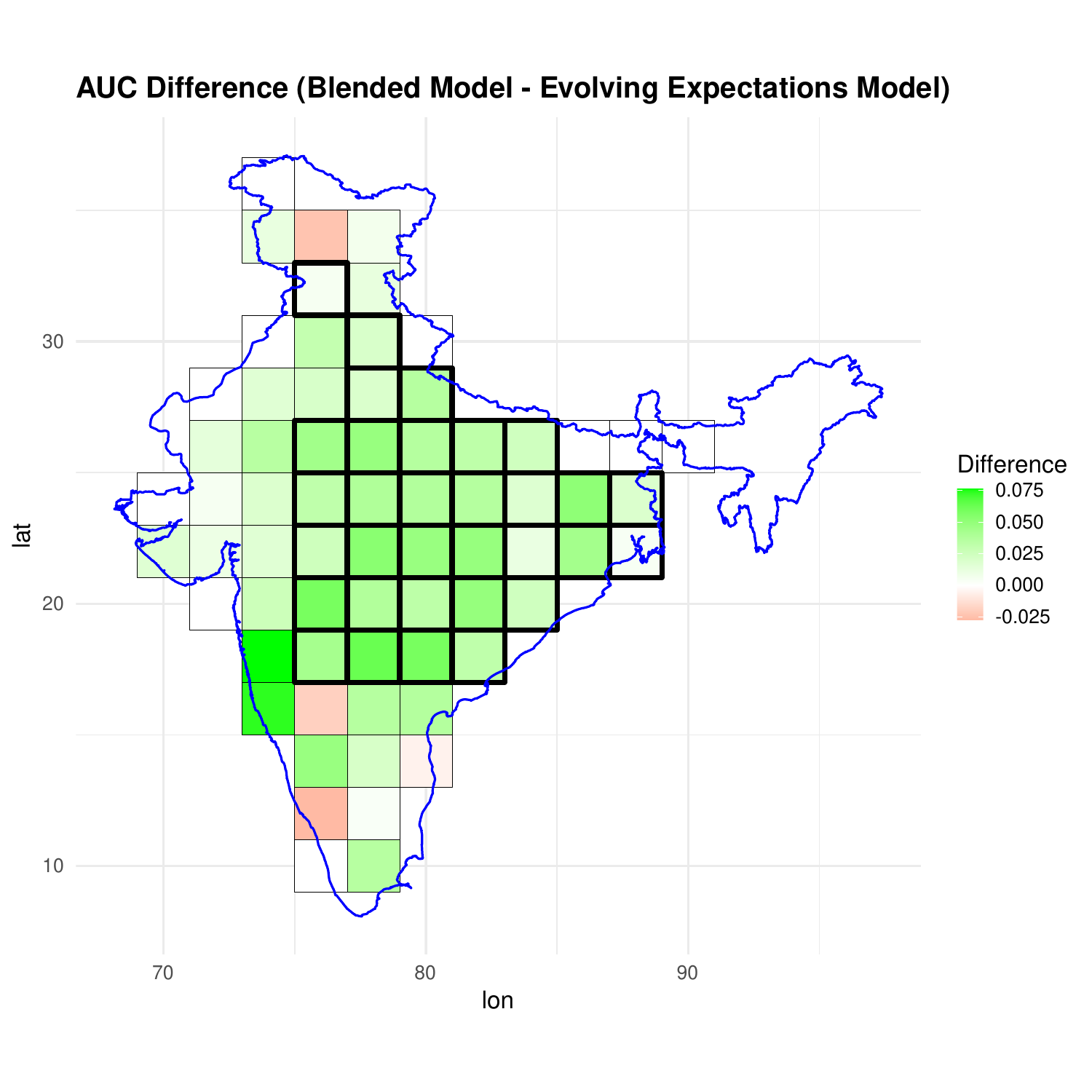}
  \caption{Map of differences between the AUC of the blended model and that of the evolving expectations model by grid cell during the 2000-2024 cross-validation period. Grid cells used for model training and intended for dissemination are depicted with thick outlines. }
\end{figure}

\begin{figure}[htbp]
  \centering
  \includegraphics[width=\textwidth]{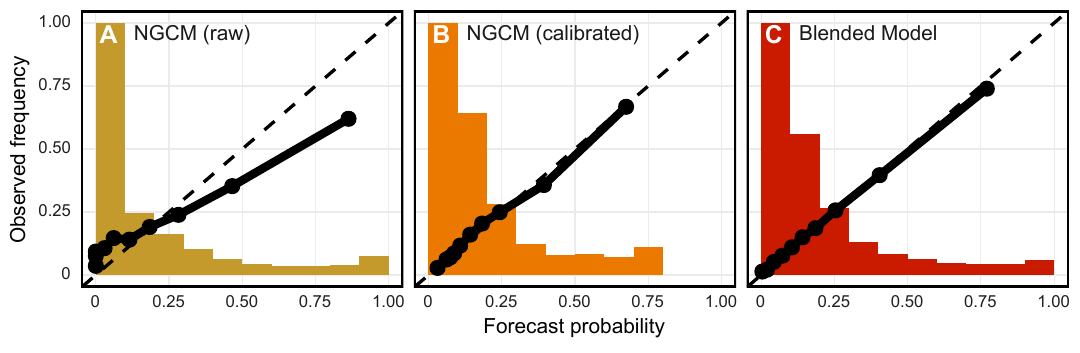}
  \caption{\textbf{Reliability diagram and histogram of probabilities assigned by different models during the 2000-2024 cross-validation period, removing the MOK filter from the onset definition.} If the model is well-calibrated the points should line up with the dashed line.  Each point represents a decile of probabilities assigned by the model, and compares the average predicted probability in that decile and the fraction of events which occurred in observation. Histograms are normalized so that the bin capturing probabilities between 0 and 10\% has height equal to one.}
\end{figure}

\begin{figure}[htbp]
  \centering
  \includegraphics[width=\textwidth]{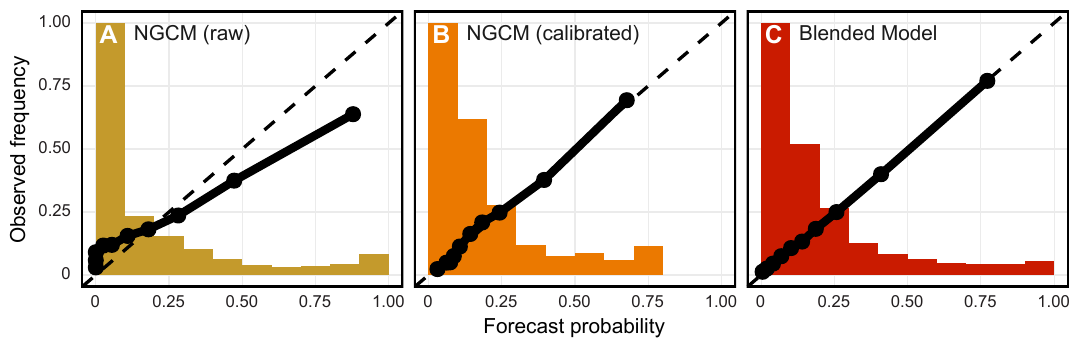}
  \caption{\textbf{Reliability diagram and histogram of probabilities assigned by different models during the 1965-1978 pre-satellite-era period, removing the MOK filter from the onset definition.} If the model is well-calibrated the points should line up with the dashed line.  Each point represents a decile of probabilities assigned by the model, and compares the average predicted probability in that decile and the fraction of events which occurred in observation. Histograms are normalized so that the bin capturing probabilities between 0 and 10\% has height equal to one.}
\end{figure}

\begin{figure}[htbp]
  \centering
  \includegraphics[width=\textwidth]{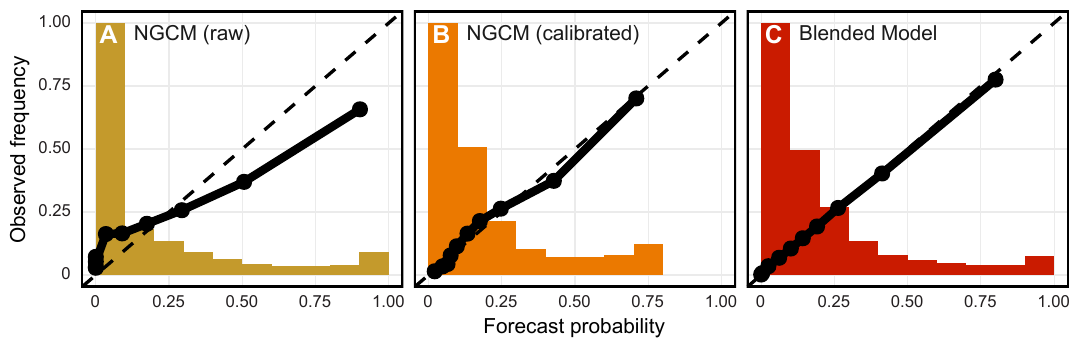}
  \caption{\textbf{Reliability diagram and histogram of probabilities assigned by different models during the 2000-2024 cross-validation period, using a climatological MOK date as part of the onset definition.} If the model is well-calibrated the points should line up with the dashed line.  Each point represents a decile of probabilities assigned by the model, and compares the average predicted probability in that decile and the fraction of events which occurred in observation. Histograms are normalized so that the bin capturing probabilities between 0 and 10\% has height equal to one.}
\end{figure}

\begin{figure}[htbp]
  \centering
  \includegraphics[width=\textwidth]{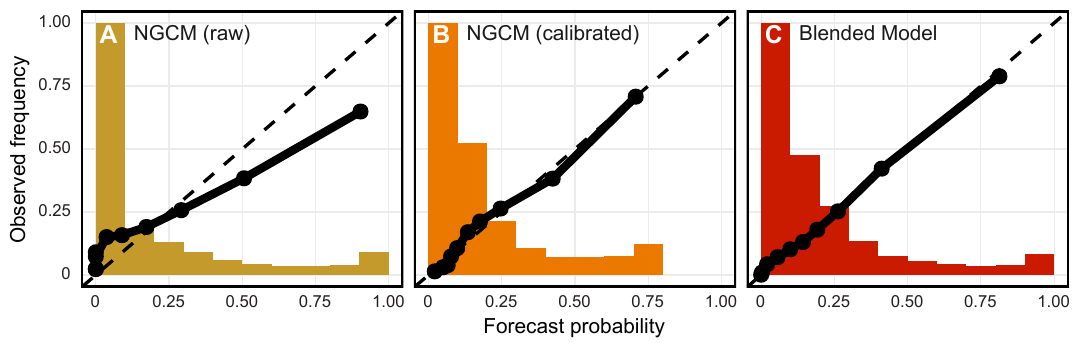}
  \caption{\textbf{Reliability diagram and histogram of probabilities assigned by different models during the 1965-1978 pre-satellite-era period, using a climatological MOK date as part of the onset definition.} If the model is well-calibrated the points should line up with the dashed line.  Each point represents a decile of probabilities assigned by the model, and compares the average predicted probability in that decile and the fraction of events which occurred in observation. Histograms are normalized so that the bin capturing probabilities between 0 and 10\% has height equal to one.}
\end{figure}

\begin{figure}[htbp]
  \centering
  \includegraphics[width=\textwidth]{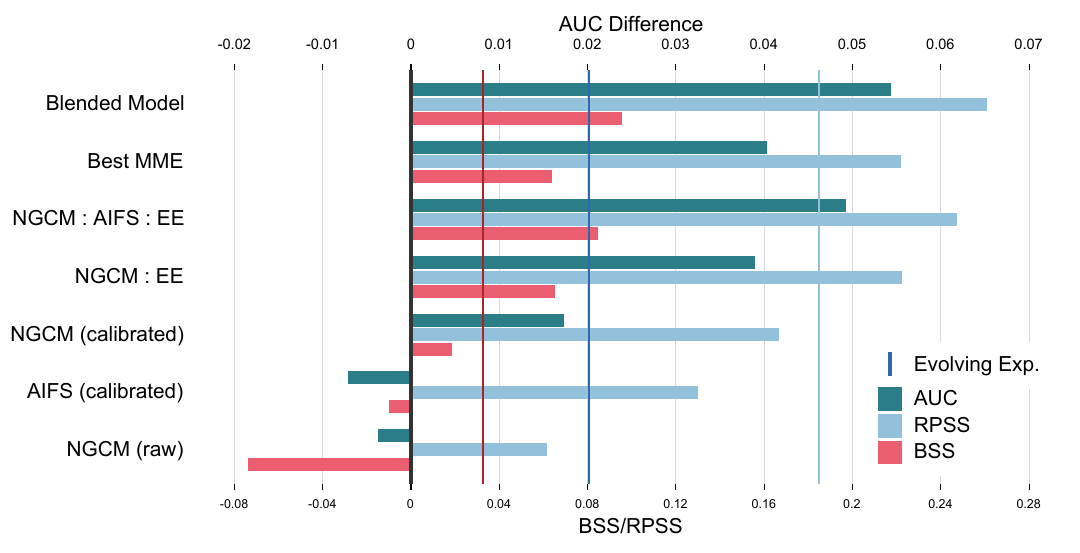}
  \caption{\textbf{Performance of models during the 2000-2024 cross-validation period, removing the MOK filter from the onset definition.} Vertical lines represent the evolving expectations model, which our decision-theory model implies should be a baseline for dissemination. The interaction models, indicated with colons, are submodels of the final blended model using 5-day rainfall variables from one or more AIWP models but not 10-day rainfall variables. The best multimodel ensemble is selected as a linear combination of probabilities from AIFS, NGCM, and the evolving expectations model maximizing the Ranked Probability Skill Score post-hoc on the 2000-2024 period. All other models (including calibration) are trained via leave-one-year-out cross-validation. Skill scores and differences in AUC are computed relative to unconditional climatology, which has an AUC of .790, a Brier Score of .612, and a Ranked Probability Score of .682. }
\end{figure}

\begin{figure}[htbp]
  \centering
  \includegraphics[width=\textwidth]{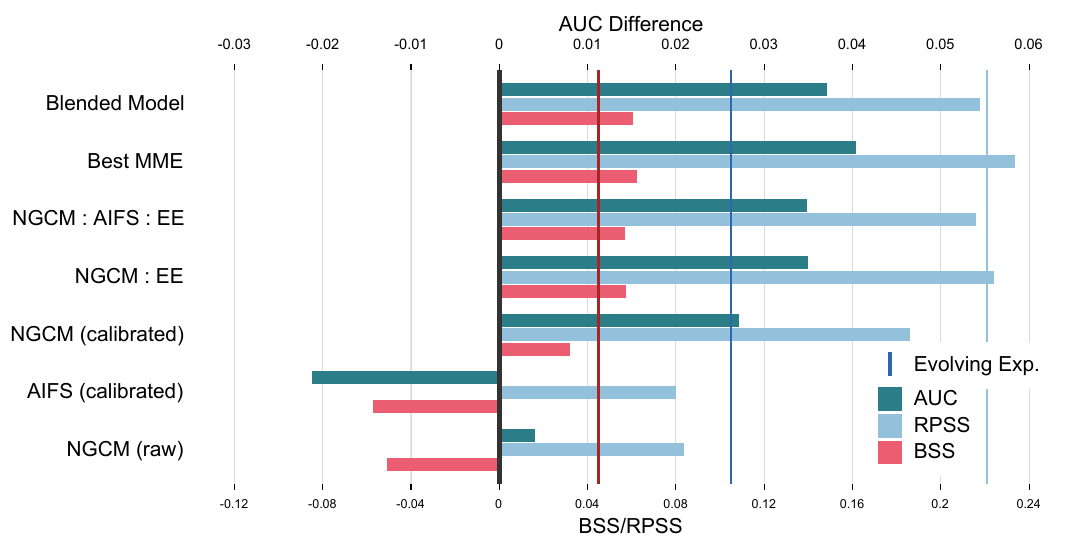}
  \caption{\textbf{Performance of models during the 1965-1978 pre-satellite-era period, removing the MOK filter from the onset definition.} Vertical lines represent the evolving expectations model, which our decision-theory model implies should be a baseline for dissemination. The interaction models, indicated with colons, are submodels of the final blended model using 5-day rainfall variables from one or more AIWP models but not 10-day rainfall variables. The best multimodel ensemble is selected as a linear combination of probabilities from AIFS, NGCM, and the evolving expectations model maximizing the Ranked Probability Skill Score post-hoc on the 2000-2024 period. All other models (including calibration) are trained via leave-one-year-out cross-validation. Skill scores and differences in AUC are computed relative to unconditional climatology, which has an AUC of .807, a Brier Score of .582, and a Ranked Probability Score of .646. }
\end{figure}

\begin{figure}[htbp]
  \centering
  \includegraphics[width=\textwidth]{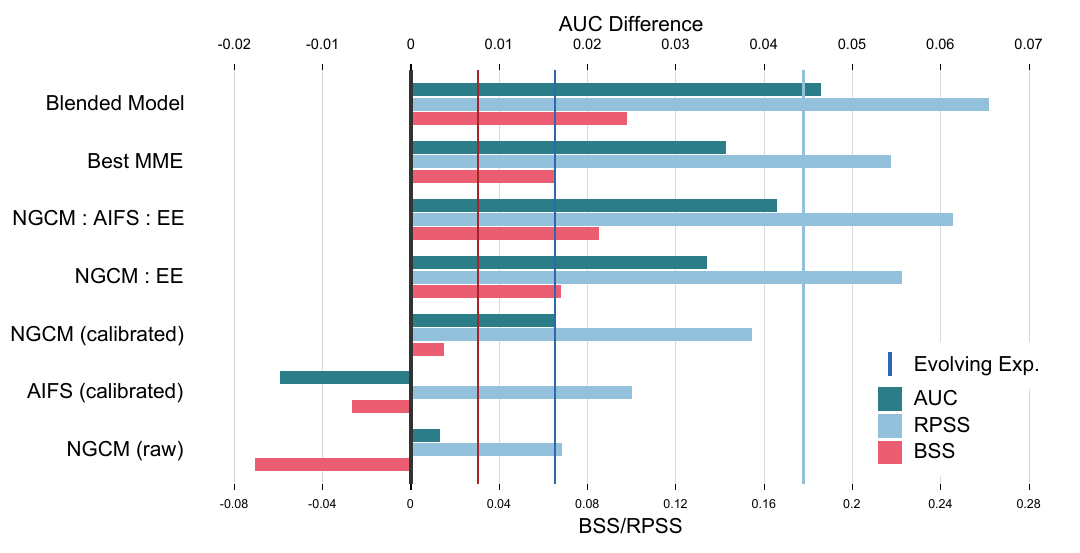}
  \caption{\textbf{Performance of models during the 2000-2024 cross-validation period, using a climatological MOK date as part of the onset definition.} Vertical lines represent the evolving expectations model, which our decision-theory model implies should be a baseline for dissemination. The interaction models, indicated with colons, are submodels of the final blended model using 5-day rainfall variables from one or more AIWP models but not 10-day rainfall variables. The best multimodel ensemble is selected as a linear combination of probabilities from AIFS, NGCM, and the evolving expectations model maximizing the Ranked Probability Skill Score post-hoc on the 2000-2024 period. All other models (including calibration) are trained via leave-one-year-out cross-validation. Skill scores and differences in AUC are computed relative to unconditional climatology, which has an AUC of .821, a Brier Score of .574, and a Ranked Probability Score of .595. }
\end{figure}

\begin{figure}[htbp]
  \centering
  \includegraphics[width=\textwidth]{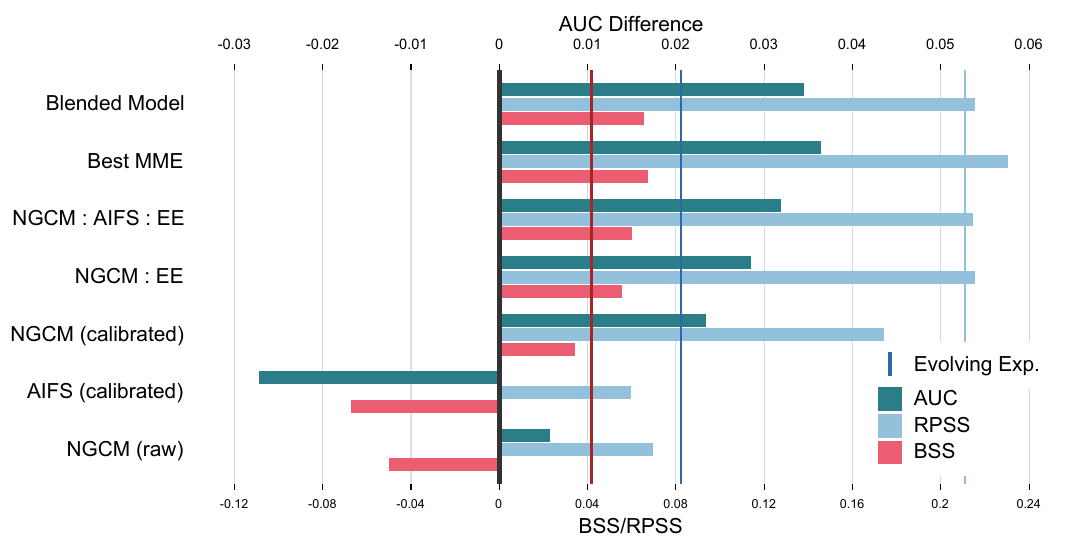}
  \caption{\textbf{Performance of models during the 1965-1978 pre-satellite-era period, using a climatological MOK date as part of the onset definition.} Vertical lines represent the evolving expectations model, which our decision-theory model implies should be a baseline for dissemination. The interaction models, indicated with colons, are submodels of the final blended model using 5-day rainfall variables from one or more AIWP models but not 10-day rainfall variables. The best multimodel ensemble is selected as a linear combination of probabilities from AIFS, NGCM, and the evolving expectations model maximizing the Ranked Probability Skill Score post-hoc on the 2000-2024 period. All other models (including calibration) are trained via leave-one-year-out cross-validation. Skill scores and differences in AUC are computed relative to unconditional climatology, which has an AUC of .827, a Brier Score of .561, and a Ranked Probability Score of .590. }
\end{figure}

\begin{figure}[htbp]
  \centering
  \includegraphics[width=\textwidth]{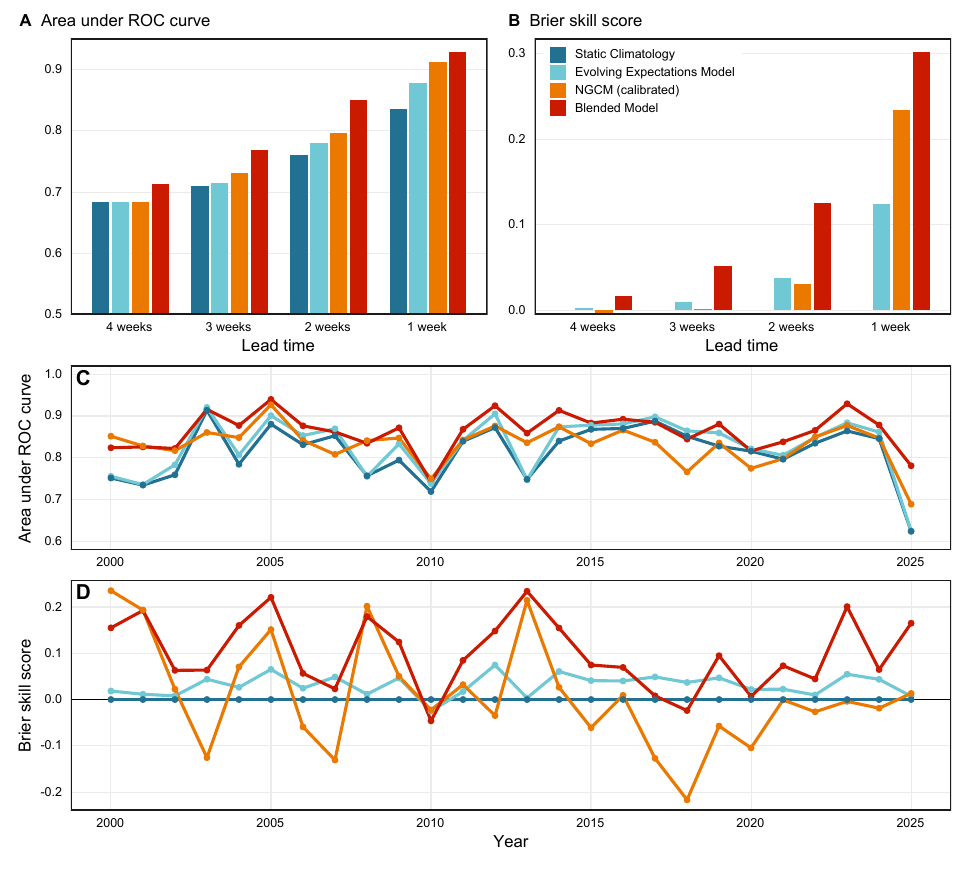}\caption{\textbf{Temporal components of model skill,  using a climatological MOK date as part of the onset definition.} \emph{(A)} Area under ROC curve (AUC) by lead time during the 2000-2024 cross-validation period. The baseline for the bars is chosen to be 0.5, indicating the AUC of a forecast with no ability to distinguish onsets from non-onsets. (B) Brier skill score by lead time, computed relative to a traditional (static) climatology model. (C) Area under ROC curve by year across all lead times. The 2000-2024 scores are computed via cross-validation. The 2025 scores are only for forecasts that were actually disseminated before MR onset in each grid cell. Dissemination began in late May, so the set of initialization dates is smaller than in other years. (D) Brier skill score by year across all lead times, calculated relative to a traditional (static) climatology model. } 
\end{figure}

\begin{figure}[htbp]
  \centering
  \includegraphics[width=\textwidth]{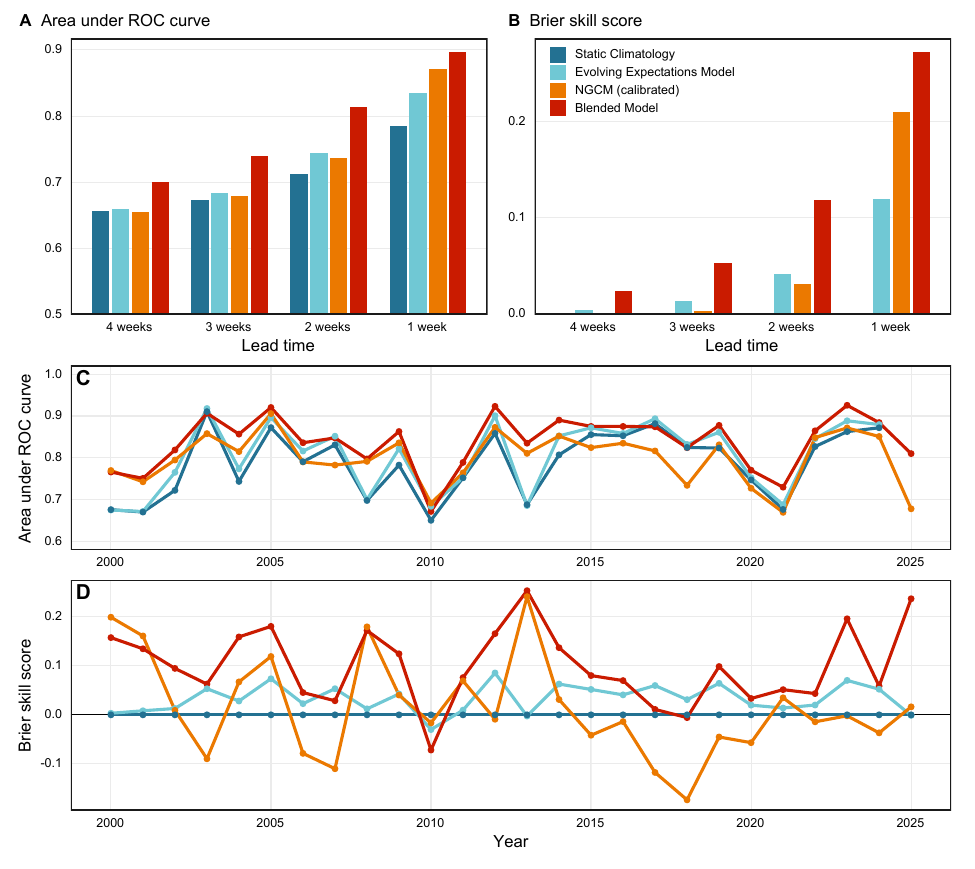}\caption{\textbf{Temporal components of model skill, removing the MOK filter from the onset definition.} \emph{(A)} Area under ROC curve (AUC) by lead time during the 2000-2024 cross-validation period. The baseline for the bars is chosen to be 0.5, indicating the AUC of a forecast with no ability to distinguish onsets from non-onsets. (B) Brier skill score by lead time, computed relative to a traditional (static) climatology model. (C) Area under ROC curve by year across all lead times. The 2000-2024 scores are computed via cross-validation. The 2025 scores are only for forecasts that were actually disseminated before MR onset in each grid cell. Dissemination began in late May, so the set of initialization dates is smaller than in other years. (D) Brier skill score by year across all lead times, calculated relative to a traditional (static) climatology model. } 
\end{figure}
\begin{figure}[htbp]
  \centering
  \includegraphics[width=\textwidth]{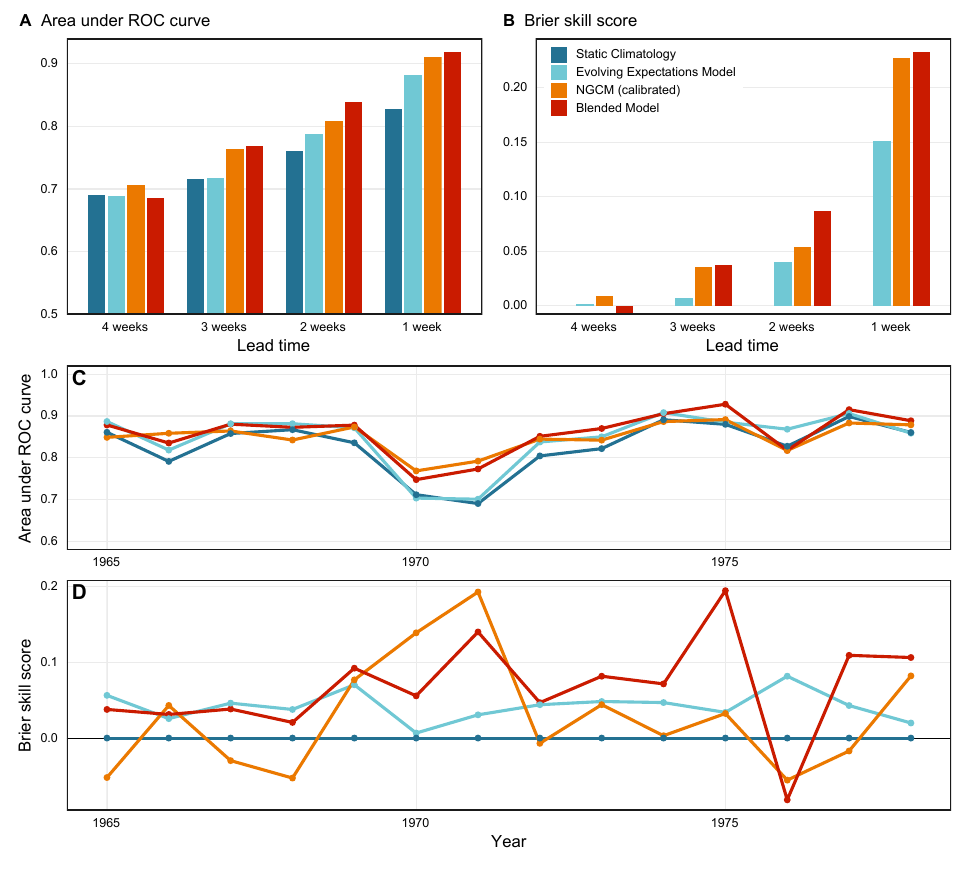}\caption{\textbf{Temporal components of model skill, using a climatological MOK date as part of the onset definition, 1965-1978.} \emph{(A)} Area under ROC curve (AUC) by lead time during the 1965-1978 pre-satellite-era period. The baseline for the bars is chosen to be 0.5, indicating the AUC of a forecast with no ability to distinguish onsets from non-onsets. (B) Brier skill score by lead time, computed relative to a traditional (static) climatology model. (C) Area under ROC curve by year across all lead times. (D) Brier skill score by year across all lead times, calculated relative to a traditional (static) climatology model. } 
\end{figure}

\begin{figure}[htbp]
  \centering
  \includegraphics[width=\textwidth]{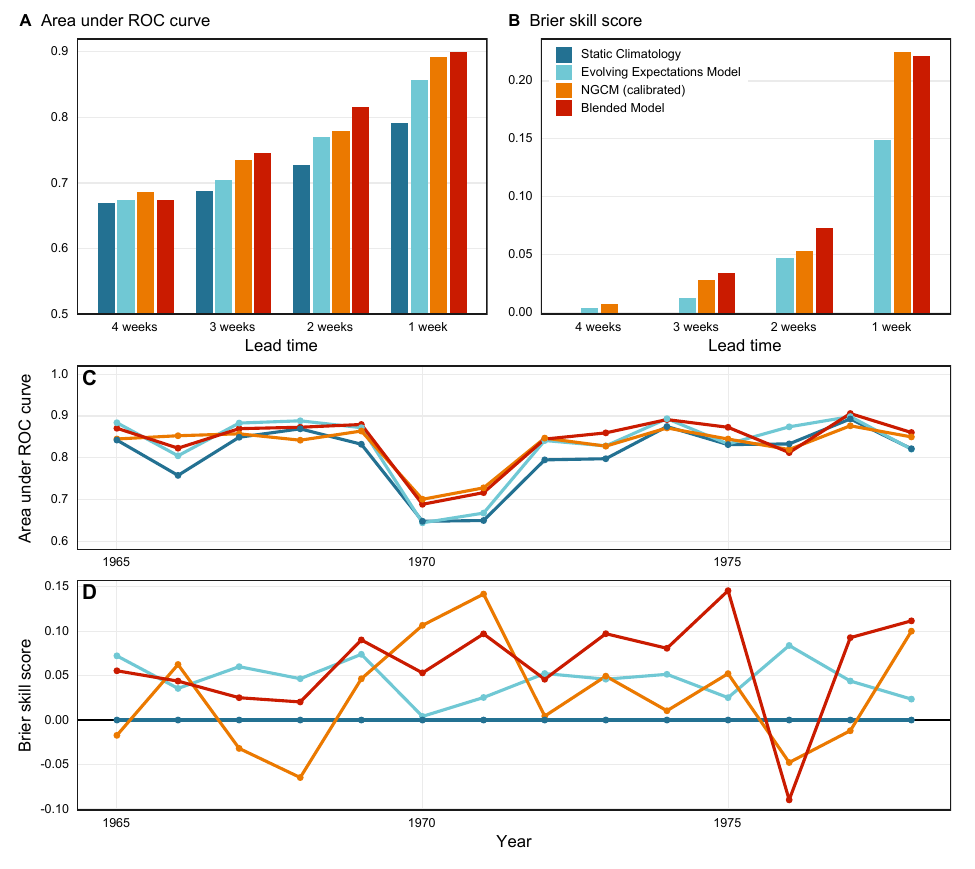}\caption{\textbf{Temporal components of model skill, removing the MOK filter from the onset definition, 1965-1978.} \emph{(A)} Area under ROC curve (AUC) by lead time during the 1965-1978 pre-satellite-era period. The baseline for the bars is chosen to be 0.5, indicating the AUC of a forecast with no ability to distinguish onsets from non-onsets. (B) Brier skill score by lead time, computed relative to a traditional (static) climatology model. (C) Area under ROC curve by year across all lead times. (D) Brier skill score by year across all lead times, calculated relative to a traditional (static) climatology model. } 
\end{figure}

\begin{figure}[htbp]
  \centering
  \includegraphics[width=\textwidth]{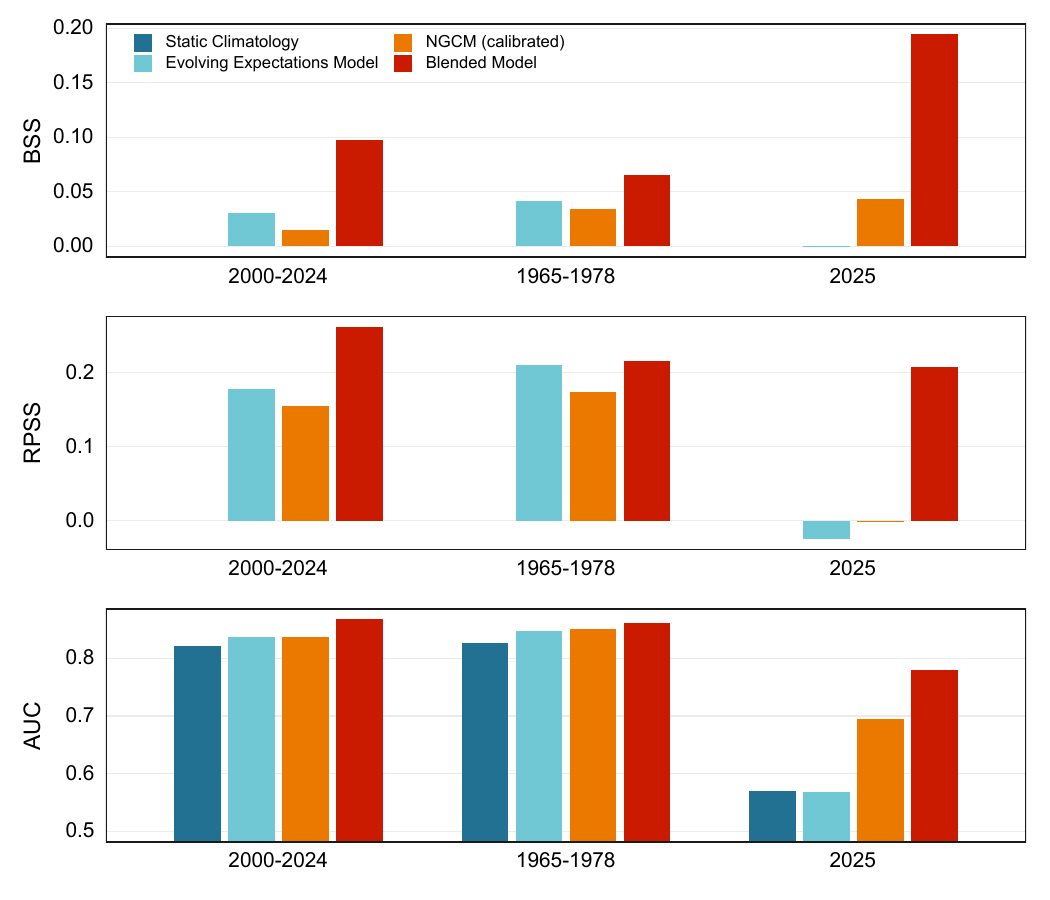}
\caption{\textbf{Skill scores aggregated by test period, using a climatological MOK date as part of the onset definition.} The blended model is cross-validated during the 2000-2024 period, and trained on 2000-2024 data during the other periods. The climatology model and the evolving expectations model are cross-validated by year using 1900-2024 IMD data. Skill scores are all computed relative to static climatology.} 
\end{figure}

\begin{figure}[htbp]
  \centering
  \includegraphics[width=\textwidth]{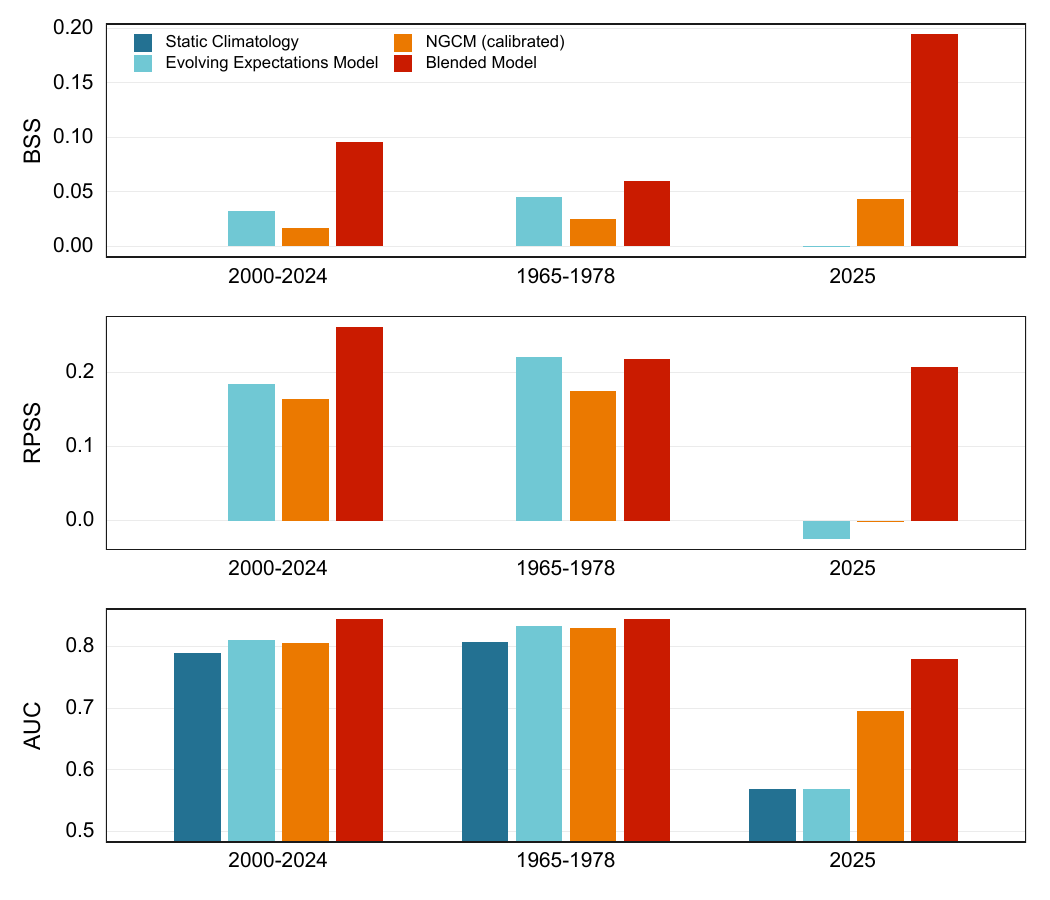}
\caption{\textbf{Skill scores aggregated by test period, removing the MOK filter from the onset definition.} The blended model is cross-validated during the 2000-2024 period, and trained on 2000-2024 data during the other periods. The climatology model and the evolving expectations model are cross-validated by year using 1900-2024 IMD data. Skill scores are all computed relative to static climatology.} 
\end{figure}

\clearpage
\bibliographystyle{plain}

\bibliography{blend_refs}